\documentclass[10pt,twocolumn,letterpaper]{article}

\usepackage[pagenumbers]{cvpr} 

\usepackage{subcaption}
\usepackage{graphicx}
\usepackage{amssymb}
\usepackage{pifont}
\usepackage{booktabs}
\usepackage{threeparttable}
\usepackage[table]{xcolor}
\usepackage[dvipsnames]{xcolor}
\definecolor{lightgreen}{RGB}{217, 234, 211}
\definecolor{lightred}{RGB}{244, 204, 204}
\definecolor{lightblue}{RGB}{201, 218, 248}
\definecolor{lightyellow}{RGB}{231, 184, 40}
\definecolor{lightblue2}{RGB}{60, 120, 216}

\definecolor{cvprblue}{rgb}{0.21,0.49,0.74}
\usepackage[pagebackref,breaklinks,colorlinks,allcolors=cvprblue]{hyperref}

\def\paperID{*****} 
\def\confName{CVPR}
\def\confYear{2026}

\title{NeighborMAE: Exploiting Spatial Dependencies between Neighboring Earth Observation Images in Masked Autoencoders Pretraining}

\author{Liang Zeng\\
KU Leuven\\
{\tt\small liang.zeng@kuleuven.be}
\and
Valerio Marsocci\\
ESA $\Phi$-lab\\
{\tt\small Valerio.Marsocci@esa.int}
\and
Wufan Zhao \\
HKUST(GZ)\\
{\tt\small wufanzhao@hkust-gz.edu.cn}
\and
Andrea Nascetti\\
KTH\\
{\tt\small nascetti@kth.se}
\and
Maarten Vergauwen\\
KU Leuven\\
{\tt\small maarten.vergauwen@kuleuven.be}
}

\begin{document}
\maketitle
\begin{abstract}
Masked Image Modeling has been one of the most popular self-supervised learning paradigms to learn representations from large-scale, unlabeled Earth Observation images. While incorporating multi-modal and multi-temporal Earth Observation data into Masked Image Modeling has been widely explored, the spatial dependencies between images captured from neighboring areas remains largely overlooked. Since the Earth's surface is continuous, neighboring images are highly related and offer rich contextual information for self-supervised learning. To close this gap, we propose NeighborMAE, which learns spatial dependencies by joint reconstruction of neighboring Earth Observation images. To ensure that the reconstruction remains challenging, we leverage a heuristic strategy to dynamically adjust the mask ratio and the pixel-level loss weight. Experimental results across various pretraining datasets and downstream tasks show that NeighborMAE significantly outperforms existing baselines, underscoring the value of neighboring images in Masked Image Modeling for Earth Observation and the efficacy of our designs. Our code is available on {\footnotesize \url{https://github.com/LeungTsang/NeighborMAE}.}
\end{abstract}    
\section{Introduction}

Self-supervised learning (SSL)~\cite{DBLP:conf/icml/ChenK0H20, DBLP:conf/cvpr/He0WXG20, DBLP:conf/cvpr/HeCXLDG22, DBLP:conf/cvpr/Xie00LBYD022} has emerged as a powerful paradigm in Earth Observation (EO), enabling the extraction of rich and transferable representations from vast amounts of unlabeled satellite and aerial imagery~\cite{DBLP:journals/corr/abs-2211-07044, DBLP:journals/corr/abs-2501-08111, DBLP:conf/nips/StewartLCWCBSR023, DBLP:journals/staeors/LongXLYYZZL21}. Such representations have proven beneficial across many downstream applications, including environmental monitoring~\cite{DBLP:journals/corr/abs-2011-05479, DBLP:conf/nips/EttenH20}, land cover classification~\cite{DBLP:conf/cvpr/TokerKWECHHSDCM22, DBLP:conf/iccv/GarnotL21, FBP2023}, and disaster management~\cite{DBLP:conf/dicta/ShenSWK23, DBLP:conf/cvpr/BonafiliaTAI20, HLS_Foundation_2023}.

\begin{figure}
\centering
\includegraphics[width=0.95\linewidth]{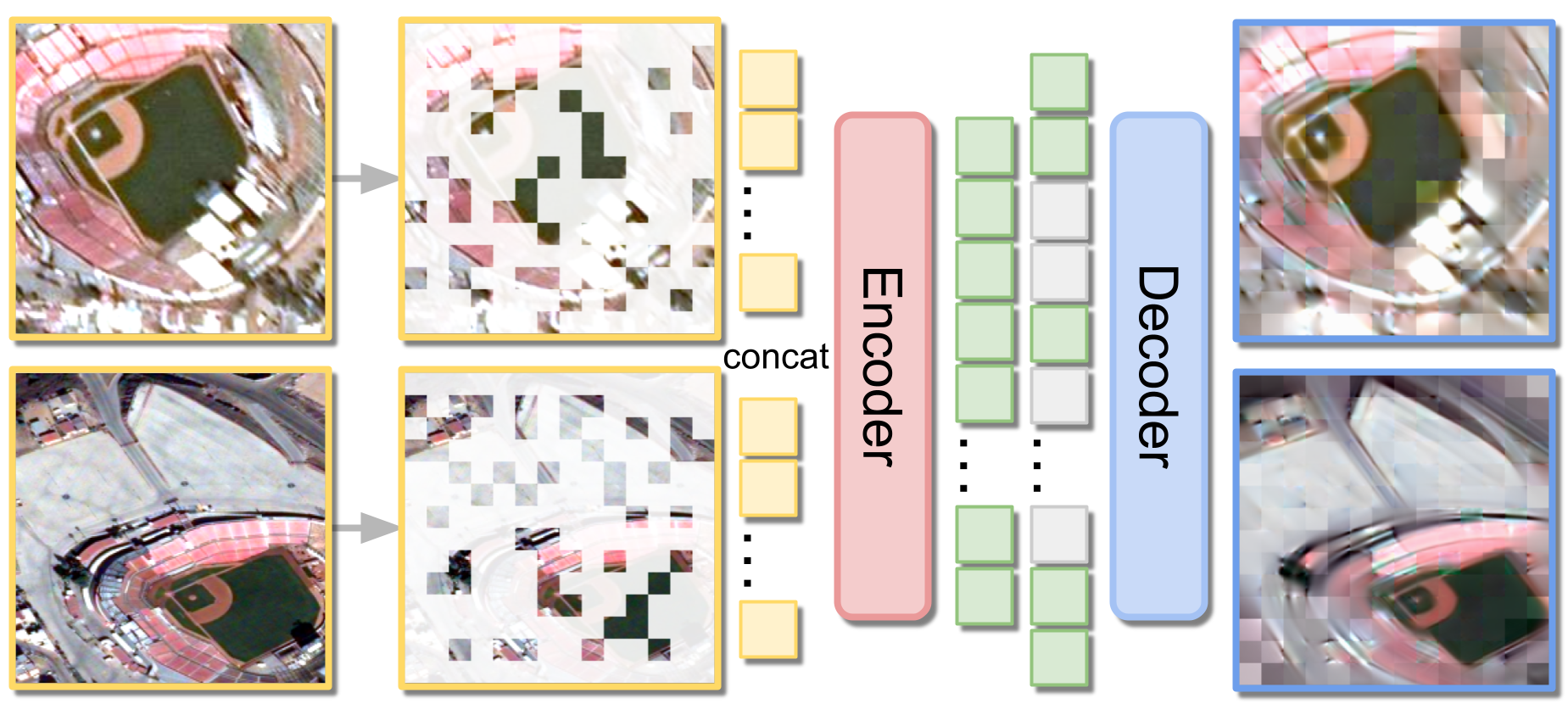}
\caption{NeighborMAE jointly reconstructs masked regions across neighboring EO images by self-attention over all tokens. This design enables the model to capture spatial and other inherent dependencies (e.g., temporal) between neighboring observations.}
\label{intro}
\end{figure}

Among recent SSL advances in EO, Masked Image Modeling (MIM)~\cite{DBLP:conf/cvpr/HeCXLDG22, DBLP:conf/cvpr/Xie00LBYD022} has proven particularly effective~\cite{DBLP:conf/nips/FullerMG23, DBLP:journals/corr/abs-2403-15356, DBLP:conf/cvpr/AstrucGML25, DBLP:journals/corr/abs-2504-11171}. By reconstructing masked patches of images, MIM-based methods compel models to learn rich contextual relationships among visible regions. Their success has been extended to multi-modal, multi-spectral, and multi-temporal EO imagery~\cite{DBLP:journals/corr/abs-2211-07044, DBLP:conf/nips/StewartLCWCBSR023, DBLP:conf/cvpr/ChristieFWM18, DBLP:journals/corr/abs-2501-08111}, underscoring the central role of mining contextual dependencies in EO data for MIM representation learning ~\cite{DBLP:conf/nips/CongKMLRHBLE22, DBLP:journals/corr/abs-2310-18660, DBLP:conf/cvpr/AstrucGML25}.

However, while EO data inherently capture the continuous and structured nature of the Earth’s surface, existing MIM frameworks are typically applied to individual image tiles, overlooking the rich spatial dependencies across neighboring scenes. This simplification restricts the learned representations to the local extent of a single image rather than modeling the broader spatial continuity that characterizes the Earth’s surface. In essence, current MIM approaches treat EO data as isolated samples, even though each image is part of a large, spatially coherent mosaic.

To address this limitation, we propose NeighborMAE, a novel MIM-based framework that explicitly models spatial dependencies across neighboring EO images. NeighborMAE jointly reconstructs masked regions from adjacent yet non-identical areas, enabling the model to learn spatial correlations through self-attention over all tokens from neighboring images. Such neighboring pairs are abundant in EO archives—arising naturally from satellite revisits, overlapping acquisitions from different missions, or spatial augmentations. They contain meaningful shared information about terrain structure, land-use continuity, and human-made infrastructure. Despite their strong spatial correlation, neighboring images have been largely overlooked in MIM-based SSL, even though they are commonly used as positive pairs in contrastive learning~\cite{DBLP:conf/iccv/AyushUMTBLE21, DBLP:conf/aaai/JeanWSALE19}.

Importantly, learning spatial dependencies from neighboring EO images is not equivalent to increasing the input image size of a static spatial pattern. Neighboring views may differ in acquisition time, viewing geometry, or sensor type, introducing different variability. Modeling these relationships encourages models to build representations that are spatially consistent yet robust to such variations — a key to generalization in real-world EO applications.

We design a principled adaptation of MAE~\cite{DBLP:conf/cvpr/HeCXLDG22} for this purpose. NeighborMAE samples neighboring image pairs based on the Intersection-over-Union (IoU) of their geospatial footprints, embeds their relative positions within a shared coordinate system, and employs dynamic masking and heuristic loss weighting to prevent shortcut learning from overlapping visible regions. Our experiments focus on RGB imagery to isolate the effect of spatial dependencies without confounding spectral factors. This also ensures fair comparison with prior MIM-based EO works. Extension to multi-spectral data is left for future work.


In summary, our main contributions are:

\begin{itemize}
\item We argue that neighboring EO images encode crucial spatial dependencies that are largely neglected in existing MIM-based SSL frameworks.
\item We propose NeighborMAE, a novel MIM framework that jointly reconstructs neighboring images to learn spatially aware representations, incorporating adaptive masking and loss weighting to prevent shortcut learning.
\item We pretrain NeighborMAE on datasets with diverse spatial and temporal distributions and evaluate the learned models across multiple downstream EO tasks on RGB imagery. The consistent improvements over baselines indicate that explicitly modeling spatial dependencies yields stronger and more generalizable representations.
\item We provide an extensive ablation studies to validate the effectiveness of our designs. Especially, we demonstrate that modeling spatial dependencies significantly improves representation quality, either alone or together with temporal dependencies, which highlights an overlooked dimension of SSL for EO images.
\end{itemize}
\section{Related Works}

\subsection{Masked Image Modeling for Earth Observation}

MIM is a popular SSL paradigm for visual representation learning, where models reconstruct masked regions of input images from the remaining visible context. Among the pioneering works, MAE~\cite{DBLP:conf/cvpr/HeCXLDG22} introduced a lightweight yet effective training strategy for Vision Transformers (ViT)~\cite{DBLP:conf/iclr/DosovitskiyB0WZ21}, while SimMIM~\cite{DBLP:conf/cvpr/Xie00LBYD022} extended MIM to hierarchical transformer architectures such as Swin Transformers~\cite{DBLP:conf/iccv/LiuL00W0LG21}.

In the EO domain, research has primarily focused on adapting MIM to multi-temporal, multi-spectral, and multi-modal data~\cite{DBLP:journals/corr/abs-2211-07044, DBLP:conf/nips/StewartLCWCBSR023, DBLP:conf/cvpr/ChristieFWM18, DBLP:journals/corr/abs-2501-08111}. Notable examples include SatMAE~\cite{DBLP:conf/nips/CongKMLRHBLE22}, Prithvi~\cite{DBLP:journals/corr/abs-2310-18660}, SpectralGPT~\cite{DBLP:journals/pami/HongZLLLYYLGJPGBC24}, and CROMA~\cite{DBLP:conf/nips/FullerMG23}, as well as more recent models such as DOFA~\cite{DBLP:journals/corr/abs-2403-15356}, AnySat~\cite{DBLP:conf/cvpr/AstrucGML25}, and TerraMind~\cite{DBLP:journals/corr/abs-2504-11171}. These works highlight the success of MIM in handling complex spectral and temporal dependencies in EO data.

However, spatial dependencies across neighboring images remains largely overlooked in MIM. A few approaches, such as ScaleMAE~\cite{DBLP:conf/iccv/ReedGLBFCKCUD23} and SatMAE++~\cite{DBLP:conf/cvpr/NomanNCA0K24}, partially capture spatial information through multi-scale reconstruction. Cross-Scale MAE~\cite{DBLP:conf/nips/TangCG023} uses neighboring images generated by data augmentations but only process them separately for auxiliary contrastive objectives, rather than being jointly modeled in reconstruction.

In contrast, NeighborMAE is designed to directly exploit true spatial dependencies between real neighboring EO images. Instead of relying on modality diversity or synthetic multi-scale cues, it leverages natural spatial continuity—information that is ubiquitous but underused in current MIM-based EO frameworks.

\subsection{Other SSL approaches for Earth Observation}

Beyond MIM, contrastive learning has been another dominant SSL approach for EO imagery. These methods learn representations by pulling together positive pairs and pushing apart negatives based on predefined similarity metrics~\cite{DBLP:conf/icml/ChenK0H20, DBLP:conf/cvpr/He0WXG20}. Typically, positive pairs are constructed from augmented views of the same image, which naturally introduces neighboring crops through random sampling.

Several works explicitly incorporate spatial and temporal proximity into contrastive frameworks. SeCo~\cite{DBLP:conf/iccv/ManasLG0R21}, CaCO~\cite{Mall_2023_CVPR} and GASSL~\cite{DBLP:conf/iccv/AyushUMTBLE21} treat geographically and temporally close tiles as positive samples, while Tile2Vec~\cite{DBLP:conf/aaai/JeanWSALE19} and SauMoCo~\cite{DBLP:journals/tgrs/KangFDLP21} refine the random cropping strategy to sample semantically related neighboring tiles more effectively. Other approaches include \cite{guo2024skysense, Akiva_2022_CVPR, tian2024swimdiff, marsocci2024crosssensor}.

However, contrastive methods rely heavily on predefined similarity assumptions and handcrafted augmentations, which may not always reflect real-world EO variability. Neighboring EO images can differ in semantics despite spatial proximity—for example, at boundaries between land-cover types or across sharp environmental transitions. NeighborMAE overcomes these limitations by adopting a reconstruction objective, allowing it to learn robust and context-aware representations without relying on heuristic similarity metrics and contrastive samples.

\begin{figure*}
  \centering
  \includegraphics[width=0.88\linewidth]{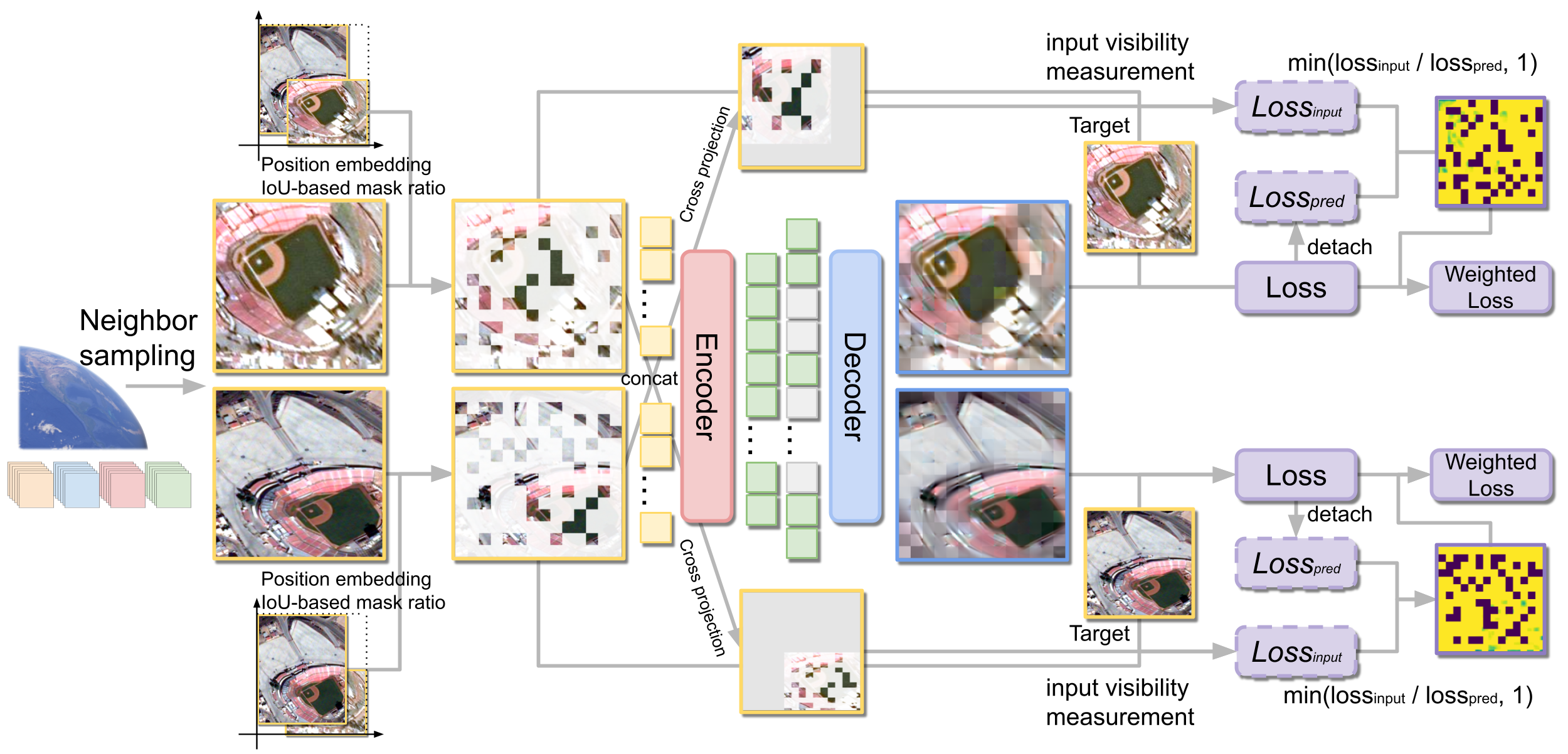}
  \caption{The overview of NeighborMAE. We sample pairs of neighboring images from datasets based on geographic coordinates. Their relative positions are embedded in a shared coordinate system, and the mask ratio is chosen based on IoU. Neighboring images are jointly reconstructed by MAE, which learns the in-between spatial dependencies by self-attention. The reconstruction loss is weighted by the visibility from the masked input of neighboring images to avoid learning shortcuts.}
  \label{overview}
\end{figure*}

\section{Methodology}

The overview of our method is shown in Figure~\ref{overview}. Our method is built upon MAE~\cite{DBLP:conf/cvpr/HeCXLDG22}, which is an SSL framework to learn visual representations by reconstructing masked portions of an input image. The MAE framework consists of two key components, an encoder for representation learning and a decoder for image reconstruction, both based on a ViT~\cite{DBLP:conf/iclr/DosovitskiyB0WZ21}. The encoder takes the visible (unmasked) patches to extract representations. The decoder receives the encoded representations and fills the missing patches with mask tokens. The mask tokens are decoded to reconstruct the missing patches in pixel space.  

Our key insight to improve MAE on EO data is to train MAE on neighboring images to learn their spatial dependencies. All visible patches from neighboring images are concatenated as input for the MAE encoder to learned joint representations. The decoder uses the joint representations to reconstruct the neighboring views collectively. Thus, models can capture the in-between spatial and temporal dependencies from neighboring images by self-attention and learn better representations. We delineate the strategy to sample neighboring images and other designs for effective learning from neighboring EO images.

\subsection{Neighboring Images Sampling}

For simplicity, we consider EO images with overlap as potential neighboring images. Given an EO dataset $D$ with metadata, we inspect the spatial coverage of all images by georeferenced bounding boxes (${\phi}_{min}$, ${\phi}_{max}$, ${\lambda}_{min}$, ${\lambda}_{max}$) and two images are considered neighboring if their Intersection over Union (IoU) is higher than a threshold $\alpha$. We pre-compute the set of neighboring images $\mathcal{N}(I_i)$ for each image $I_i$ as Eqn. ~\ref{neighbor} and save them as a lookup table to avoid online searching. During training, for each iteratively sampled image $I_i$, we randomly sample an image $I_j$ from its neighboring candidates $\mathcal{N}(I_i)$.

\begin{equation}
\mathcal{N}(I_i) = \{I_j\vert I_j \in D, IoU(I_i,I_j)>\alpha\}, I_i \in D
\label{neighbor}
\end{equation}

Apart from spatial coverage, no temporal, mission, or atmospheric constraints are imposed by the sampling of neighboring images. These additional factors (e.g., capture time, mission source, cloud coverage) are inherent properties of the used dataset, which naturally determine the degree of temporal or radiometric variation present among neighbors. Such variations also contribute useful diversity that enhances representation learning by NeighborMAE. 

\subsection{Data Augmentations}

A set of artificial augmentations $A$ is applied to the sampled raw neighboring image pair to generate the final input $((A(I_i),A(I_j)))$ for the network. Particularly, Random Crop can artificially introduce further spatial variations in addition to the original spatial distribution of the raw images in dataset. It ensures that the final input neighboring EO image pairs $((A(I_i),A(I_j)))$ have spatial variations even original $I_i,I_j$ have the same coverage (i.e. temporal revisits).

Let $i$, $j$, $h$, $w$ denote the parameters (left-top corner and crop size) used in Random Crop and $H$, $W$ denote the original image size, the new georeferenced bounding boxes (${\phi}^{'}_{min}$, ${\phi}^{'}_{max}$, ${\lambda}^{'}_{min}$, ${\lambda}^{'}_{max}$) after Random Crop are computed according to Eqn. \ref{bbox} (corner cases are omitted for simplicity). After cropping, the images are resized to the model input size $H^{'}$, $W^{'}$.

\begin{equation}
\begin{array}{l}
{\phi}^{'}_{min} = {\phi}_{min} - ({\phi}_{max} - {\phi}_{min})\frac{i+h}{H} \\
{\phi}^{'}_{max} = {\phi}_{min} - ({\phi}_{max} - {\phi}_{min})\frac{i}{H} \\
{\lambda}^{'}_{min} = {\lambda}_{min} + ({\lambda}_{max} - {\lambda}_{min})\frac{j}{W} \\
{\lambda}^{'}_{max} = {\lambda}_{min} + ({\lambda}_{max} - {\lambda}_{min})\frac{j+w}{W}
\end{array}
\label{bbox}
\end{equation}

\subsection{Relative Positional Embedding}
To allow effective learning of spatial dependencies between neighboring images, we must embed their fine-grained relative positions in a shared coordinate system. To guarantee that the positional embedding remains consistent in downstream applications without geographic metadata, we normalize the georeferenced bounding boxes to $[0, 1]$ by the minimum and maximum latitudinal and longitudinal coordinates of a neighboring image pair by Eqn. \ref{norm} (corner cases are omitted for simplicity). Such relative positional embedding can be flexibly computed in the image coordinate system and do not require absolute positions and distances on the Earth's surface. For example, for single-image input, the normalized bounding box is simply constant (0, 1, 0, 1).

\begin{equation}
\begin{array}{l}
top_i = \frac{{\phi}^{'i}_{max} - min({\phi}^{'i}_{min}, {\phi}^{'j}_{min})}{max({\phi}^{'i}_{max}, {\phi}^{'j}_{max}) - min({\phi}^{'i}_{min}, {\phi}^{'j}_{min})} \\
bottom_i = \frac{{\phi}^{'i}_{min} - min({\phi}^{'i}_{min}, {\phi}^{'j}_{min})}{max({\phi}^{'i}_{max}, {\phi}^{'j}_{max}) - min({\phi}^{'i}_{min}, {\phi}^{'j}_{min})} \\
left_i = \frac{{\lambda}^{'i}_{min} - min({\lambda}^{'i}_{min}, {\lambda}^{'j}_{min})}{max({\lambda}^{'i}_{max}, {\lambda}^{'j}_{max}) - min({\lambda}^{'i}_{min}, {\lambda}^{'j}_{min})} \\
right_i = \frac{{\lambda}^{'i}_{max} - min({\lambda}^{'i}_{min}, {\lambda}^{'j}_{min})}{max({\lambda}^{'i}_{max}, {\lambda}^{'j}_{max}) - min({\lambda}^{'i}_{min}, {\lambda}^{'j}_{min})}
\end{array}
\label{norm}
\end{equation}

The patch-level bounding boxes are then computed accordingly after the neighboring images are evenly patchified. We use the standard sinusoidal positional encoding~\cite{DBLP:conf/nips/VaswaniSPUJGKP17} to match the 4-dimensional coordinates of bounding boxes to the dimension of the used vision transformer. Finally, a learnable image-level embedding is added to the patch-level relative positional embedding to discriminate tokens from either image.

\subsection{Dynamic Mask Ratio}
Neighboring images may make the learning task easier since additional information is provided. Especially, neighboring images can be partially seen as sequences and MAE-based methods applied to videos~\cite{DBLP:conf/nips/TongS0022, DBLP:conf/cvpr/WangHZTHWWQ23} usually require a higher mask ratio. Thus, we introduce a dynamic mask ratio to adjust the difficulty of the reconstruction task based on the IoU of the \textbf{augmented} neighboring image pairs $(A(I_i), A(I_j))$, which can be computed by Eqn. \ref{mask_ratio} given the lower bound $m_1$ and the upper bound $m_2$. We compensate for the learning difficulty with a higher mask ratio when reconstructing neighboring images with more overlap.

\begin{equation}
\begin{array}{l}
mask\ ratio = m_1 + IoU(A(I_i), A(I_j)) * (m_2 - m_1) \\
\end{array}
\label{mask_ratio}
\end{equation}

\subsection{Weighted Loss by Input Visibility}
Visible patches are ignored in MAE reconstruction to ensure models to only focus on the challenging reconstruction of invisible patches. Since we introduce neighboring images into the learning, we categorize pixels to be reconstructed into three cases after masking: visible from the original image (\textit{self visible}), its location is visible from the corresponding position of its neighboring image (\textit{cross visible}), and not visible from both neighboring images (\textit{not visible}). The joint reconstruction of neighboring images may introduce a learning shortcut when a pixel is \textit{cross visible} without a significant change, so that the models can simply learn to copy-paste it from the neighboring image as a prediction. We quantify the input visibility for \textit{cross visible} pixels by the loss of using its neighboring correspondence as prediction, and the reconstruction loss for \textit{cross visible} is bounded by this value.

To establish the positional correspondences between pixels from $I_i^{'}$ and $I_j^{'}$ and identify different categories of pixels, we use the transform matrix $T_i$ from the image coordinate system of $I_i^{'}$ to the shared coordinate system, and transform matrix $T^{-1}_j$ from the shared coordinate system to the image coordinate system of $I_j^{'}$, as Eqn. ~\ref{titj}. Given the homogeneous coordinates $(u, v, 1)$ of a pixel $p_i$ from $I_i^{'}$, its corresponding pixel $p_i^j$ in $I_j^{'}$ can be computed by Eqn. ~\ref{tjtip}

\begin{equation}
\begin{aligned}
T_i= \left[
\begin{array}{ccc}
\frac{{right_i}-{left_i}}{W^{'}} & 0 & left_i \\
0 & \frac{{bottom_i}-{top_i}}{H^{'}} & top_i \\
0 & 0 & 1
\end{array}
\right] \\
T_j^{-1}= \left[
\begin{array}{ccc}
\frac{W^{'}}{{right_j}-{left_j}} & 0 & \frac{W^{'}\times left_j}{{left_j}-{right_j}} \\
0 & \frac{H^{'}}{{bottom_j}-{top_j}} & \frac{H^{'}\times top_j}{{top_j}-{bottom_j}} \\
0 & 0 & 1
\end{array}
\right]
\end{aligned}
\label{titj}
\end{equation}


\begin{equation}
p_i^j= T^{-1}_j T_i p_i
\label{tjtip}
\end{equation}

Given the pair of input augmented neighboring images $(I_i^{'}, I_j^{'})$, the binary random masks $(M_i, M_j)$, and the reconstructed images $(R_i, R_j)$, \textit{self visible}, \textit{cross visible}, and \textit{not visible} pixels from $I_i^{'}$ can be identified by Eqn. \ref{pixel}.

\begin{equation}
\begin{array}{l}
P_{self} = \{p_i \vert \neg M_i(p_i),\ p_i \in I_i^{'}\}  \\
P_{cross} = \{p_i \vert \neg M_j(p_i^j) \wedge M_i(p_i),\ p_i \in I_i^{'},\ p_i^j \in I_j^{'}\} \\
P_{not} = \{p_i \vert M_j(p_i^j) \wedge  M_i(p_i),\ p_i \in I_i^{'},\ p_i^j \in I_j^{'}\} \\
\end{array}
\label{pixel}
\end{equation}

We use Mean Squared Error (MSE) loss for the image reconstruction, and the loss for a \textit{cross visible} pixel is bounded by the loss directly using its visible correspondences from the neighboring image as prediction using the weight as Eqn. ~\ref{loss_weight}. Note that the weight value is detached from the gradient computation graph.

\begin{equation}
weight = \left\{
\begin{array}{cl}
0 &  p_i \in P_{self} \\
min(\frac{MSE(I^{'}_j(p_i^j), I^{'}_i(p_i)))}{MSE(R_i(p_i), I^{'}_i(p_i))}, 1) & p_i \in P_{cross} \\
1 &  p_i \in P_{not}  \\
\end{array} \right.
\label{loss_weight}
\end{equation}
\section{Experiments}

\begin{table*}\small
  \centering
  \begin{threeparttable}[b]
  \begin{tabular}{@{}c|c|c|c|c|c|c|c|c @{}}
    \toprule
           &                &   fMoW      & UC Merced   & RESISC45    &  FireRisk  &  ForestNet  &  FBP      &  PASTIS     \\
    Method &           Data &  Acc.~\cite{DBLP:conf/cvpr/ChristieFWM18}       & Acc.~\cite{DBLP:conf/gis/YangN10}        & Acc.~\cite{DBLP:journals/pieee/ChengHL17}        &  Acc.~\cite{DBLP:conf/dicta/ShenSWK23}       &  Acc.~\cite{DBLP:journals/corr/abs-2011-05479}       &  mIoU~\cite{FBP2023}       &  mIoU ~\cite{DBLP:conf/eccv/AstrucGML24}      \\
    \midrule
    SatMAE~\cite{DBLP:conf/nips/CongKMLRHBLE22}           & fMoW~\cite{DBLP:conf/cvpr/ChristieFWM18} & 64.3 / 76.9 & 86.9 / 93.1 & 86.4 / 95.6 & 60.0 / 63.1 & 44.7 / 54.8 & 51.2 / 56.8 & 30.1 / 32.4 \\
    Temporal SatMAE~\cite{DBLP:conf/nips/CongKMLRHBLE22}  & fMoW~\cite{DBLP:conf/cvpr/ChristieFWM18} & 57.6 / 77.6 & 79.7 / 94.2 & 82.0 / 94.5 & 60.1 / 63.3 & 43.8 / 52.4 & 50.9 / 56.4 & 31.7 / 33.9 \\
    ScaleMAE~\cite{DBLP:conf/iccv/ReedGLBFCKCUD23}         & fMoW~\cite{DBLP:conf/cvpr/ChristieFWM18} & \underline{67.1} / \underline{78.8} & \underline{93.1} / 95.9 & \underline{92.2} / \underline{96.6} & \underline{61.3} / \underline{64.0} & 50.8 / 56.3 & 57.5 / 65.7 & 31.8 / 34.5 \\
    CrossScale MAE~\cite{DBLP:conf/nips/TangCG023}   & fMoW~\cite{DBLP:conf/cvpr/ChristieFWM18} & 42.6 / 71.4 & 73.8 / 93.1 & 87.2 / 91.1 & 58.2 / 61.6 & 41.1 / 49.7 & 41.6 / 47.5 & 28.8 / 31.4 \\
    SatMAE++~\cite{DBLP:conf/cvpr/NomanNCA0K24}         & fMoW~\cite{DBLP:conf/cvpr/ChristieFWM18} & 66.4 / 77.4 & 90.9 / 97.1 & 86.4 / 96.2 & 61.0 / 63.9 & 48.3 / 54.2 & 54.4 / 61.9 & 31.5 / 34.2 \\
    MAE~\cite{DBLP:conf/cvpr/HeCXLDG22}              & fMoW~\cite{DBLP:conf/cvpr/ChristieFWM18} & 66.8 / 78.2 & 87.8 / 94.5 & 89.8 / 96.0 & 60.7 / 63.5 & 50.1 / 55.5 & 58.8 / 63.9 & 31.1 / 33.9 \\ 
    \rowcolor{lightgreen}
    NeighborMAE      & fMoW~\cite{DBLP:conf/cvpr/ChristieFWM18} & \textbf{68.8} / \textbf{79.3} & 91.4 / \underline{97.6} & 91.0 / \underline{96.6} & \textbf{61.4} / \textbf{64.2} & \textbf{52.4} / \textbf{57.0} & \textbf{60.3} / \textbf{66.6} & 31.8 / \textbf{36.1}  \\ 
    \midrule
    MAE~\cite{DBLP:conf/cvpr/HeCXLDG22}              & Satl.~\cite{DBLP:journals/corr/abs-2501-08111} & 57.3 / 76.7 & 88.1 / 95.9 & 87.7 / 95.0 & 60.1 / 63.2 & 50.4 / 56.0 & 57.0 / 62.5 & 31.2 / 34.1 \\
    \rowcolor{lightgreen}
    NeighborMAE      & Satl.~\cite{DBLP:journals/corr/abs-2501-08111} & 58.8 / 77.9 & 88.8 / 96.2 & 88.5 / 95.6 & 60.4 / \underline{64.0} & \underline{51.5} / \underline{56.8} & 58.4 / 63.7 & \textbf{32.5} / 35.4 \\
    \midrule
    DOFA~\cite{DBLP:journals/corr/abs-2403-15356}             & DOFA~\cite{DBLP:journals/corr/abs-2403-15356} & 62.6 / 78.0 & \textbf{96.4} / \textbf{98.3} & \textbf{94.5} / \textbf{97.4} & 60.3 / \underline{64.0} & 43.8 / 54.0 & \underline{59.7} / \underline{66.2} & \underline{32.2} / \underline{35.6} \\
    \bottomrule
   \end{tabular}
  \caption{Downstream image classification and semantic segmentation by frozen backbone (left) or full fine-tuning (right). Bold and underline mark the \textbf{best} and \underline{second best} score. NeighborMAE outperforms MAE and previous baselines in the same research line by noticeable margins while being competitive against the state-of-the-art DOFA of large-scale pretraining.}
  \label{main result}
  \end{threeparttable}
\end{table*}

\subsection{Baselines}
The objective of this paper is to study the effectiveness of learning spatial dependencies between neighboring EO images in MIM in consistent experiment settings, which is an independent factor from multi-spectral and multi-modal EO images. To this end, we follow the research line of SatMAE~\cite{DBLP:conf/nips/CongKMLRHBLE22}, Scale MAE~\cite{DBLP:conf/iccv/ReedGLBFCKCUD23}, Cross-Scale MAE~\cite{DBLP:conf/nips/TangCG023}, SatMAE++~\cite{DBLP:conf/cvpr/NomanNCA0K24} (temporal and non-temporal) because we share the same idea of learning spatial (scale) and temporal dependencies between EO images and limit the scope of experiments to RGB-based models and tasks for simplicity. Additionally, we also reproduce vanilla MAE~\cite{DBLP:conf/cvpr/HeCXLDG22} as a plain baseline and include DOFA~\cite{DBLP:journals/corr/abs-2403-15356} as a strong baseline, which is one of the state-of-the-art models trained on large-scale multi-modal, multi-spectral datasets and readily compatible with RGB-only experiments in our setting. All models are based on ViT-Large-16 using the model weights trained by the authors, except for MAE. Potential adaptations are described in section ~\ref{sec:baseline} of the supplementary materials.

\subsection{Pretraining Datasets}

Following SatMAE~\cite{DBLP:conf/nips/CongKMLRHBLE22}, Scale MAE~\cite{DBLP:conf/iccv/ReedGLBFCKCUD23}, Cross-Scale MAE~\cite{DBLP:conf/nips/TangCG023}, and SatMAE++~\cite{DBLP:conf/cvpr/NomanNCA0K24}, we conduct NeighborMAE pretraining on the fMoW-RGB dataset~\cite{DBLP:conf/cvpr/ChristieFWM18}, which has 363,572 images from 83,412 multi-temporal sequences. Each image represents a specific semantic class, so the sizes of images in fMoW can vary significantly. The IoU threshold $\alpha$ to sample neighboring images is empirically set to 0.1 on fMoW, to exclude the case that an image is only a tiny crop of another with weak spatial dependencies. In addition to the curated, object-centric fMoW dataset, we perform experiments on the newly-released Satellogic dataset (Satl.) from EarthView~\cite{DBLP:journals/corr/abs-2501-08111}. Satellogic is generated from 3,758 captures, and each capture spans over a large area. 6,165,992 image patches of $384\times384$ are automatically cropped from these captures without the awareness of semantics by a sliding window. Only 986,521 patches have at least two revisits. Therefore, Satellogic challenges self-supervised methods by incomplete objects and limited context in a single image, and less temporal dependencies from the dataset. We use the RGB channels of Satellogic-RGB for our experiments. The IoU threshold $\alpha$ is set to 0.0 on Satellogic, so tightly adjacent patches are treated as neighboring images. We show example neighboring images in section ~\ref{sec:morevisualization} of the supplementary materials.

\subsection{Implementation Details}

We follow previous works to train ViT-Large-16 for 800 epochs on fMoW. For a similar training length, the models are trained for 50 epochs on Satellogic. The batch size is set to 2048, and the learning rate is given by $1.5e^{-4} \times bs / 256$. The lower bound $m_1$ and upper bound $m_2$ of the mask ratio are set to 0.75 and 0.85, respectively. All other settings are kept the same as the original MAE~\cite{DBLP:conf/cvpr/HeCXLDG22}, referring to section~\ref{sec:pretraining} in the supplementary materials.

To maintain the equivalent computation regarding epoch and batch size as general MAE, the number of epochs for NeighborMAE is defined by the same number of images as the dataset size that has been used for training. The batch size is defined by the actual number of images in a batch instead of the number of image pairs.

\subsection{Evaluation Protocols}
The pretrained models are evaluated by transfer learning performance to image classification and semantic segmentation, as frozen and tunable backbones. The implementation is based on the codebase of the original MAE~\cite{DBLP:conf/cvpr/HeCXLDG22} for classification and the UperNet~\cite{DBLP:conf/eccv/XiaoLZJS18} with a ViT backbone in MMSegmentation~\cite{mmseg2020} for segmentation. To match the pretraining setting of NeighborMAE and the baselines, we consider a wide range of downstream benchmarks using RGB remote sensing images, including fMoW-RGB~\cite{DBLP:conf/cvpr/ChristieFWM18}, UC Merced~\cite{DBLP:conf/gis/YangN10}, RESISC-45~\cite{DBLP:journals/pieee/ChengHL17}, FireRisk~\cite{DBLP:conf/dicta/ShenSWK23}, ForestNet~\cite{DBLP:journals/corr/abs-2011-05479} image classification and Five-Billion-Pixels~\cite{FBP2023}, PASTIS-HD~\cite{DBLP:conf/eccv/AstrucGML24} semantic segmentation. The full settings are given in section ~\ref{sec:eval} in the supplementary materials.

\textbf{fMoW-RGB}~\cite{DBLP:conf/cvpr/ChristieFWM18} is a variant of the fMoW-RGB dataset for classification of functional purposes of buildings and land use, which is generated by cropping the original images by the given bounding boxes. Models are trained for 20 epochs with AdamW ($lr = 2e^{-3}$, cosine annealing), batch size 512, and input size 224. We use the official train/val split with 363572 and 53041 images, respectively.

\textbf{UC Merced}~\cite{DBLP:conf/gis/YangN10} is a land use classification image dataset. Models are trained for 20 epochs with AdamW ($lr = 2.5e^{-4}$, cosine annealing), batch size 64, and input size 224. We use the TorchGeo split~\cite{Stewart_TorchGeo_Deep_Learning_2022} with 1260 images for training and 420 images for validation.

\textbf{RESISC45}~\cite{DBLP:journals/pieee/ChengHL17} is a remote sensing scene classification dataset. Models are trained for 20 epochs with AdamW ($lr = 2.5e^{-4}$, cosine annealing), batch size 128, and input size 224. We use the TorchGeo split~\cite{Stewart_TorchGeo_Deep_Learning_2022} with 18900 for training and 6300 images for validation.

\textbf{FireRisk}~\cite{DBLP:conf/dicta/ShenSWK23} is a remote sensing dataset for fire risk
assessment. Models are trained for 20 epochs with AdamW ($lr = 1e^{-3}$, cosine annealing), batch size 256, and input size 224. We use the official train/val split with 70331 and 21541 images, respectively.

\textbf{ForestNet}~\cite{DBLP:journals/corr/abs-2011-05479} is a dataset to classify the drivers of deforestation. Models are trained for 20 epochs with AdamW ($lr = 2.5e^{-4}$, cosine annealing), batch size 64, and input size 224. We use the official train/val split with 4435 and 1236 images, respectively.

\textbf{Five-Billion-Pixels (FBP)}~\cite{FBP2023} is a land cover semantic segmentation dataset of rich categories, large coverage, wide distribution, and high spatial resolution. We train models for 20000 iterations with AdamW ($lr = 1e^{-4}$, cosine annealing), batch size 16, and crop size 768. We use its official RGB version and split for experiments with 120 for training and 30 images for validation.

\textbf{PASTIS-HD}~\cite{DBLP:conf/eccv/AstrucGML24} is a variant of PASTIS dataset~\cite{DBLP:conf/iccv/GarnotL21} for segmentation of agricultural parcels using high-resolution RGB images. Models are trained for 20000 iterations with AdamW ($lr = 1e^{-4}$, cosine annealing), batch size 16, and crop size 512. We use the fold 0-3 with 1937 images for training and fold 4 with 496 images for validation.

\subsection{Main Results}

Table ~\ref{main result} presents the transfer learning performance of NeighborMAE and we discuss the results compared to different baselines in the following paragraphs. We report the average performance of 3 independent runs.

\paragraph{Comparison with MAE} NeighborMAE consistently obtains better performance than its immediate baseline of MAE. For example, NeighborMAE increases the linear probing accuracy on fMoW classification by $+2.0\%$ and $1.5\%$, and the fine-tuning accuracy by $+1.1\%$ and $1.2\%$ when pretraining on fMoW-RGB and Satellogic-RGB, respectively. For semantic segmentation, fine-tuning fMoW-pretrained NeighborMAE yields $+2.7\%$ mIoU on Five-Billion-Pixels and $+2.2\%$ mIoU on PASTIS-HD. The results indicate that neighboring images introduce rich data dependencies, and the designs of NeighborMAE enable effective learning from neighboring images based on multi-temporal sequences of selected areas of interest (fMoW-RGB) or patches automatically cropped by a sliding window from captures with fewer revisits (Satellogic-RGB).

\paragraph{Comparison within fMoW pretraining} Among existing models pretrained on fMoW-RGB, NeighborMAE achieves the best performance on the in-domain fMoW classification, indicating that more semantic cues are learned from fMoW by NeighborMAE than previous baselines. NeighborMAE is also one of the best-performing methods on UC Merced and RESISC45. For specific applications of fire risk assessment and deforestation monitoring, NeighborMAE also surpasses previous models by significant margins. 

\paragraph{Comparison with DOFA} Based on our evaluation protocol on RGB tasks, NeighborMAE gains overall competitive performance against DOFA~\cite{DBLP:journals/corr/abs-2403-15356}, one of the state-of-the-art models pretrained on large-scale multi-modal, multi-spectral datasets. DOFA indeed demonstrates strong performance on land use and scene classification, including UC Merced and RESISC45. Over other downstream tasks, NeighborMAE pretrained on satellogic can match DOFA and fMoW-pretrained NeighborMAE even slightly outperforms DOFA. The results suggest that the current RGB-based NeighborMAE is on par with the state-of-the-art models on RGB downstream tasks. Its future extension to multi-modal, multi-spectral data is promising.
\section{Ablation Studies}

To understand the contributions of each key component in NeighborMAE, we conduct ablation studies using ViT-Base models pretrained on fMoW~\cite{DBLP:conf/cvpr/ChristieFWM18} dataset for 200 epochs or Satellogic~\cite{DBLP:journals/corr/abs-2501-08111} for 10 epochs, keeping all other settings consistent with the main experiments. The models are transferred to fMoW classification~\cite{DBLP:conf/cvpr/ChristieFWM18} and FBP semantic segmentation~\cite{FBP2023} and we report the average performance of 3 independent runs.

\subsection{Source of Performance Gain}
\begin{figure}[h]
  \centering
  \includegraphics[width=0.99\linewidth]{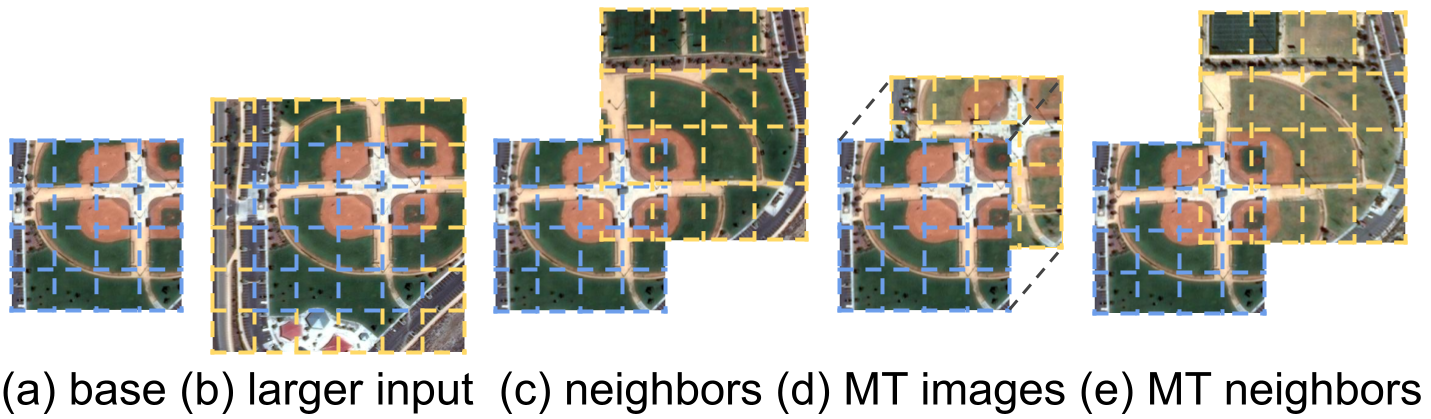}
  \caption{Examples of different ways to extend the view of a \textcolor{lightblue2}{base input} (a) using \textcolor{lightyellow}{a similar number of additional patches}. Neighbors introduce diverse spatial variations for learning.}
  \label{ab_dependencies_fig}
\end{figure}

We conduct an experiment to verify that our innovation of learning spatial dependencies from neighboring EO images is responsible for the observed performance gain and compare neighbors to other means to extend the input views in MAE pretraining. Figure ~\ref{ab_dependencies_fig} illustrates different types of extended input using a similar number of additional patches and Table ~\ref{ab_dependencies}. Simply enlarging the input area (b) to 320 yields a fixed spatial layout with uniform resolution and limited performance improvement. In contrast, sampling neighboring images (c) provides richer spatial patterns and patches with various resolutions and geometric distortions, which can boost the learning as the experiment suggests. When co-located temporal images are introduced in (d), similar performance gain to uni-temporal neighbors are observed. Without additionally increasing the token budget, multi-temporal neighboring images (e) further vary the spatial coverage and demonstrate significantly better representation quality in all downstream evaluations, likely due to the positive interaction of spatial and temporal variations for learning.



\begin{table}[h]
  \centering
  \begin{tabular}{@{}c|l|c|c @{}}
    \toprule
               &               &   fMoW~\cite{DBLP:conf/cvpr/ChristieFWM18}      &   FBP~\cite{FBP2023}  \\
    Data       & Input type   &       Acc.     &   mIoU    \\
    \midrule
    fMoW~\cite{DBLP:conf/cvpr/ChristieFWM18}       & (a) \textit{base}    & 58.0 / 76.5 &  53.5 / 58.2 \\
    fMoW~\cite{DBLP:conf/cvpr/ChristieFWM18}       & (b) \textit{larger size}  & 58.2 / 76.7  & 53.8 / 58.6  \\
    fMoW~\cite{DBLP:conf/cvpr/ChristieFWM18}       & (c) \textit{neighbors}  & 58.7 / 76.8 &  54.1 / 58.9 \\
    fMoW~\cite{DBLP:conf/cvpr/ChristieFWM18}       & (d) \textit{MT images}  & 58.2 / 77.0 &  54.5 / 58.7 \\
    fMoW~\cite{DBLP:conf/cvpr/ChristieFWM18}       & (e) \textit{MT neighbors}  & \textbf{61.7} / \textbf{77.7} &  \textbf{56.0} / \textbf{60.4} \\
    \midrule
    Satl.~\cite{DBLP:journals/corr/abs-2501-08111}       & (a) \textit{base}    & 49.5 / 74.3 &  51.1 / 55.5 \\
    Satl.~\cite{DBLP:journals/corr/abs-2501-08111}       & (b) \textit{larger size}  & 49.6 / 74.3 &  51.3 / 55.8  \\
    Satl.~\cite{DBLP:journals/corr/abs-2501-08111}       & (c) \textit{neighbors}    & 50.1 / 75.0 &  51.6 / 56.0 \\
    Satl.~\cite{DBLP:journals/corr/abs-2501-08111}       & (d) \textit{MT images}   & 50.6 / 74.7 &  52.4 / 56.4 \\
    Satl.~\cite{DBLP:journals/corr/abs-2501-08111}       & (e) \textit{MT neighbors}   & \textbf{52.4} / \textbf{75.8} &  \textbf{53.2} / \textbf{56.8} \\
    \bottomrule
  \end{tabular}
  \caption{Using the same input length budget ($2\times$ base), pretraining on neighboring images achieved the best performance in both uni-temproal or multi-temporal setting. Moreover, the joint learning spatial and temporal dependencies provides synergistic benefits.}
  \label{ab_dependencies}
\end{table}

\subsection{Dynamic Mask Ratio}
We study the best range of the dynamic mask ratio based on the IoU of the input neighboring images by pretraining on fMoW-RGB. From Table \ref{ab_mask_ratio}, our proposed dynamic mask ratio outperforms a constant mask ratio of 0.75 or 0.80. The optimal lower bound of the mask ratio is kept the same at 0.75 as MAE when two neighboring images have little overlap. The optimal upper bound of the mask ratio is achieved at 0.85 when two neighboring images cover the same area. 

\begin{table}[h]
  \centering
  \begin{tabular}{@{}c|c|c|c @{}}
    \toprule
                &        &   fMoW~\cite{DBLP:conf/cvpr/ChristieFWM18}       &   FBP~\cite{FBP2023}     \\
      $m_1$     & $m_2$  &    Acc.      &   mIoU      \\
    \midrule
        0.75    &  0.75  &  61.0 / 77.4 & 55.0 / 59.8 \\
        0.75    &  0.80  &  61.5 / 77.6 & 55.7 / 60.3 \\
        0.75    &  0.85  &  \textbf{61.7} / \textbf{77.7} & \textbf{56.0} / \textbf{60.4} \\
        0.75    &  0.90  &  61.4 / 77.3 & 55.8 / 60.1  \\
        0.80    &  0.80  &  60.8 / 77.3 & 55.1 / 59.4  \\
        0.80    &  0.85  &  61.1 / 77.0 & 55.2 / 59.5  \\
        
    \bottomrule
  \end{tabular}
  \caption{Dynamic mask ratio based on the IoU of neighboring image pairs noticeably improve the performance with lower bound $m_1$ at 0.75 and upper bound $m_2$ at 0.85.}
  \label{ab_mask_ratio}
\end{table}

\subsection{Weighted Loss by Input Visibility}

We validate the effectiveness of our heuristic loss weight by input visibility and Table \ref{ab_loss_weight} presents the results. Compared to our method, full reconstruction, including all unmasked pixels, significantly degrades the learned representations. Excluding all self-visible and cross-visible pixels also slightly reduces performance, since the reconstruction of cross-visible pixels with changes is still meaningful. Since fMoW mainly consists of long-term multi-temporal sequences with distinct changes, the effect of our loss weighting is similar to full weighting for cross-visible pixels.  However, Satellogic provides fewer revisits and neighboring images are more frequently cropped from the same image. Our weighted loss improves its pretraining by lowering the weight for cross-visible pixels of few or no changes.

\begin{table}[h]
  \centering
  \begin{tabular}{@{}c|c|c|c|c @{}}
    \toprule
           &  self-   & cross-   &   fMoW~\cite{DBLP:conf/cvpr/ChristieFWM18}       &   FBP~\cite{FBP2023}      \\
    data   &  visible & visible  &    Acc.     &   mIoU       \\
    \midrule
    fMoW~\cite{DBLP:conf/cvpr/ChristieFWM18}    &    0.    &  ours    & \textbf{61.7} / \textbf{77.7} &  \textbf{56.0} / 60.4 \\
    fMoW~\cite{DBLP:conf/cvpr/ChristieFWM18}    &    0.    &    1.    & 61.6 / \textbf{77.7} &  55.8 / \textbf{60.5} \\
    fMoW~\cite{DBLP:conf/cvpr/ChristieFWM18}    &    0.    &    0.    & 61.4 / 77.4 &  55.7 / 60.2 \\
    fMoW~\cite{DBLP:conf/cvpr/ChristieFWM18}    &    1.    &    1.    & 58.6 / 77.0 &  55.0 / 59.1 \\
    \midrule
    Satl.~\cite{DBLP:journals/corr/abs-2501-08111}   &    0.    &  ours    & \textbf{52.4} / \textbf{75.8} &  \textbf{53.2} / \textbf{56.8} \\
    Satl.~\cite{DBLP:journals/corr/abs-2501-08111}   &    0.    &    1.    & 51.8 / 75.1 &  52.5 / 55.6 \\
    Satl.~\cite{DBLP:journals/corr/abs-2501-08111}   &    0.    &    0.    & 51.9 / 74.9 &  52.8 / 56.2 \\
    Satl.~\cite{DBLP:journals/corr/abs-2501-08111}   &    1.    &    1.    & 50.2 / 74.1 &  51.4 / 55.3 \\    
    \bottomrule
  \end{tabular}
  \caption{Although our weighted reconstruction loss does not show salient effect on fMoW pretraining with rich temporal changes, it improves pretraining on Satellogic by decreasing the weight for pixels visible from the neighboring image (cross-visible) since images from Satellogic have much fewer temporal changes.}
  \label{ab_loss_weight}
\end{table}

\section{Learning Efficiency}
We investigate the learning efficiency of MAE~\cite{DBLP:conf/cvpr/HeCXLDG22}, SatMAE++~\cite{DBLP:conf/cvpr/NomanNCA0K24} and NeighborMAE by ViT-Base backbone pretrained on fMoW-RGB~\cite{DBLP:conf/cvpr/ChristieFWM18} with a total batch size of 1024 using 4 H100 GPUs. Our reproduced MAE and SatMAE++ use the model implemented by the original authors but trained by the same training scripts as our NeighborMAE for a fair comparison.

\begin{figure}[h]
  \centering
  \begin{subfigure}{0.43\linewidth}
    \includegraphics[width=1\linewidth]{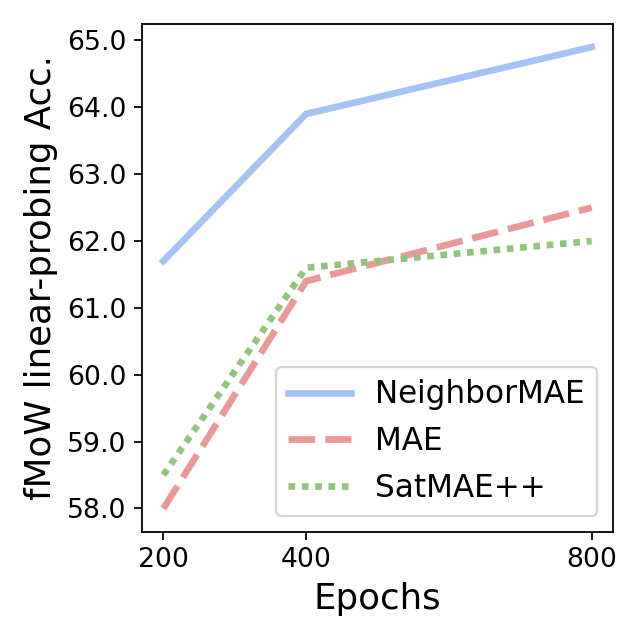}
  \end{subfigure}
  \begin{subfigure}{0.43\linewidth}
    \includegraphics[width=1\linewidth]{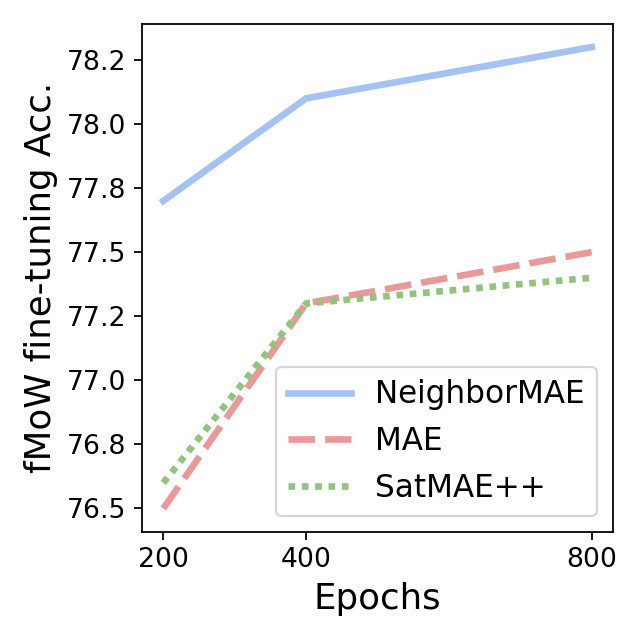}
  \end{subfigure}
  
  \caption{fMoW classification performance with different epochs pretraining on fMoW by NeighborMAE, MAE and SatMAE++. NeighborMAE consistently obtains the best performance.}
  \label{ab_efficiency_fig}
\end{figure}

\paragraph{Performance efficiency} Figure ~\ref{ab_efficiency_fig} investigates the performance efficiency on fMoW classification for different pretrained models. NeighborMAE consistently outperforms MAE and SatMAE++ with different pretraining epochs and the three methods demonstrate similar convergence speed.


\paragraph{Computation efficiency} We measure average batch time and maximum memory usage during the pretraining for each method. The joint encoding and decoding of tokens from two images in NeighborMAE leads to slightly longer batch time and more memory usage than MAE, mainly due to the $O(n^2)$ computation and space complexity of self-attention. However, NeighborMAE mitigates the downside by better performance efficiency. Both MAE and NeighborMAE are much less demanding than methods based on multi-scale reconstruction like SatMAE++, because the involved upsampling is extremely computationally expensive.

\begin{table}[h]\small
  \centering
  \begin{tabular}{@{}c|c|c @{}}
    \toprule
    method &  batch  time (s) &   memory per GPU   \\
    \midrule
    MAE          & 0.122    &   15.5 GB \\
    NeighborMAE  & 0.134    &   19.6 GB \\
    SatMAE++     & 0.381    &   58.7 GB \\
    \bottomrule
  \end{tabular}
  \caption{Batch time and memory usage when training ViT-Base with a total batch size of 1024 distributed on 4 H100 GPUs. NeighborMAE only requires slightly more computational resources than MAE, while multi-scale reconstruction used in SatMAE++ is computationally expensive due to upsampling.}
  \label{computation}
\end{table}
\section{Discussion and Conclusion}

SSL has been increasingly important for its ability to learn from large-scale unlabeled data. However, digesting massive data is inevitably expensive in computation, and the amount of data is not infinite after all. To maximize the computation and data efficiency, the inherent dependencies hidden in data must be exploited. In this work, we highlighted the importance of spatial dependencies between neighboring EO images, which is largely overlooked by existing MIM-based SSL approaches. To address this gap, we introduced NeighborMAE, a spatially aware MIM framework that jointly reconstructs neighboring EO images with adaptive masking and loss weighting to ensure a challenging pretext task for SSL. Through extensive pretraining on datasets with diverse spatial and temporal characteristics, followed by evaluations on multiple downstream EO tasks using RGB imagery, we demonstrated consistent improvements over previous baselines, revealing an underexplored yet powerful direction for SSL in EO.

Looking ahead, we aim to extend NeighborMAE to multi-spectral and multi-modal EO data for a comprehensive understanding of various EO data, building on the strong results achieved on RGB imagery. When scaling to more than two neighboring images, future work may reduce computational overhead through more efficient token-reduction strategies or next-generation architectures that bypass the $O(n^2)$ complexity of self-attention. We hope that our findings encourage further exploration of SSL tailored to the unique characteristics of EO data.

{
    \small
    \bibliographystyle{ieeenat_fullname}
    \bibliography{main}
}

\clearpage
\setcounter{page}{1}
\maketitlesupplementary

\setcounter{section}{0}


\section{Pretraining Details}
\label{sec:pretraining}

We show the pretraining hyperparameters in Table ~\ref{setting_pt}, which are based on the original MAE. The training time of the main fMoW experiments on 4 H100s is given in Table ~\ref{re_computation}.

\begin{table}[h]
  \centering
  \begin{threeparttable}
  \begin{tabular}{@{}c|c|c @{}}
    \toprule
    Pretrain Data      & fMoW         & Satellogic                             \\
    \midrule

    input size         & \multicolumn{2}{c}{224}                          \\
    base lr (batch 256)& \multicolumn{2}{c}{1.5e-4}                       \\
    actual lr          & \multicolumn{2}{c}{1.2e-3}                       \\
    lr schedule        & \multicolumn{2}{c}{cosine decay}                 \\
    weight decay       & \multicolumn{2}{c}{0.05}                         \\
    optimizer          & \multicolumn{2}{c}{AdamW}                        \\
    AdamW betas        & \multicolumn{2}{c}{$\beta_1, \beta_2=0.9, 0.95$}       \\
    batch size         & \multicolumn{2}{c}{2048}                           \\
    augmentation       &  \multicolumn{2}{c}{RandomResizedCrop}   \\
    crop scale         &  \multicolumn{2}{c}{(0.2, 1)} \\
    epochs             &    800       &    50                            \\
    warmup epochs      &    40        &    2.5                           \\
    \midrule
    IoU threshold $\alpha$  &    0.1      &    0.0 \\
    Lower mask ratio $m_1$              &   \multicolumn{2}{c}{0.75} \\
    Upper Mask ratio $m_2$              &   \multicolumn{2}{c}{0.85} \\
    
    \bottomrule
  \end{tabular}
  \caption{Pre-training settings}
  \label{setting_pt}
  \end{threeparttable}
\end{table}

\begin{table}[h]
  \centering
  \begin{tabular}{@{}c|c|c @{}}
    \toprule
    Model   & batch time & wall time \\
    \midrule
    MAE \scriptsize{-Large}          &  0.298s    & 11h 31m 16s \\
    NeighborMAE \scriptsize{-Large}  &  0.364s    & 14h 01m 50s \\
    \bottomrule
  \end{tabular}
  \caption{Training time of the main fMoW experiments on 4 H100s.}
  \label{re_computation}
\end{table}

\section{Evaluation Protocols}
\label{sec:eval}

\subsection{Image Classification Details}
\label{sec:classification}

We show the training settings for image classification in the Table ~\ref{setting_cls}, which are based on the evaluation script of the original MAE. We do not use auto augmentation and color jitter as we find them suboptimal for EO classification tasks.

\begin{table}[h]
  \centering
  \begin{threeparttable}
  \begin{tabular}{@{}c|c|c @{}}
    \toprule
    Experiment setting & linear probing & fine-tuning                    \\
    \midrule
    input size         & \multicolumn{2}{c}{224}                          \\
    base lr (batch 256)&   1e-3       &  1e-3                            \\
    layer-wise lr decay&   N/A        &   0.75                        \\
    lr schedule        & \multicolumn{2}{c}{cosine decay}                 \\
    weight decay       &    0         & 0.05                         \\
    optimizer          &  \multicolumn{2}{c}{AdamW}                        \\
    AdamW betas        &  \multicolumn{2}{c}{$\beta_1, \beta_2=0.9, 0.95$} \\
    batch size         &  \multicolumn{2}{c}{depends on datasets}  \\
    warmup epochs      &    1        &    1                              \\
    epochs             &    20       &    20                            \\
    augmentation       & \multicolumn{2}{c}{RandomResizedCrop}   \\
    crop scale         & \multicolumn{2}{c}{(0.08, 1)} \\
    label smoothing    &     0        &   0.1     \\
    mixup              &     0        &   0.8    \\
    cutmix             &     0        &   1.0    \\
    drop path          &     0        &   0.2    \\
    global pooling     &  cls token   & average  \\
    \bottomrule
  \end{tabular}
  \caption{Image classification settings used in evaluation protocols.}
  \label{setting_cls}
  \end{threeparttable}
\end{table}

\subsection{Semantic Segmentation Details}
\label{sec:segmentation}

We show the training settings for semantic segmentation in the Table ~\ref{setting_seg}, which are based on the configuration to fine-tune MAE models with UperNet~\cite{DBLP:conf/eccv/XiaoLZJS18} head in MMSegmentation~\cite{mmseg2020}.

\begin{table}[h]
  \centering
  \begin{threeparttable}
  \begin{tabular}{@{}c|c|c @{}}
    \toprule
    Experiment setting & frozen backbone & fine-tuning                    \\
    \midrule
    input size         & \multicolumn{2}{c}{depends on datasets}          \\
    base lr (batch 256)&    \multicolumn{2}{c}{1e-4}                            \\
    layer-wise lr decay&   N/A         &   0.75                        \\
    lr schedule        & \multicolumn{2}{c}{cosine decay}                 \\
    weight decay       &    0          & 0.05                         \\
    optimizer          & \multicolumn{2}{c}{AdamW}                        \\
    momentum           & \multicolumn{2}{c}{$\beta_1, \beta_2=0.9, 0.999$} \\
    batch size         & \multicolumn{2}{c}{16}  \\
    warmup iterations  & \multicolumn{2}{c}{1000}                     \\
    iterations         & \multicolumn{2}{c}{20000}    \\     
    augmentation       & \multicolumn{2}{c}{RandomScale, RandomCrop}   \\
    scale range        & \multicolumn{2}{c}{(0.5, 2.0)} \\
    layers for feature pyramid          & \multicolumn{2}{c}{7, 11, 15, 23} \\
    inference          & \multicolumn{2}{c}{sliding windows on original images} \\
    \bottomrule
  \end{tabular}
  \caption{Semantic segmentation settings used in evaluation protocol.}
  \label{setting_seg}
  \end{threeparttable}
\end{table}

\section{Adaptations for Baseline Models}
\label{sec:baseline}

We use the publicly available model weights of baseline models, except for our reproduced MAE on EO datasets. All models are based on ViT-Large-16 with potential subtle differences. We upsample the positional embedding of Cross-Scale MAE~\cite{DBLP:conf/nips/TangCG023} to an input size of 224 since it was trained on an input size 128. For the temporal variant of SatMAE~\cite{DBLP:conf/nips/CongKMLRHBLE22}, we replace its timestamp embedding with a learnable positional embedding as timestamps are unavailable in downstream tasks. We use B-G-R input according to the pretraining of SatMAE++~\cite{DBLP:conf/cvpr/NomanNCA0K24}. The input for all baseline models is normalized by the $mean$ and $std$ used in their original pretraining.

\section{Visualization}
\label{sec:morevisualization}

\begin{figure}[h]
  \centering
      \begin{subfigure}{0.18\linewidth}
  \includegraphics[width=1\linewidth]{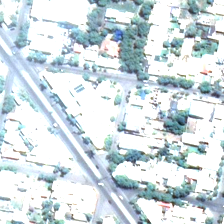}
  \end{subfigure}
  \begin{subfigure}{0.18\linewidth}
  \includegraphics[width=1\linewidth]{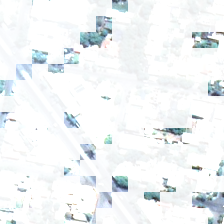}
  \end{subfigure}
  \begin{subfigure}{0.18\linewidth}
  \includegraphics[width=1\linewidth]{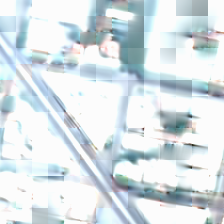}
  \end{subfigure}
  \begin{subfigure}{0.18\linewidth}
  \includegraphics[width=1\linewidth]{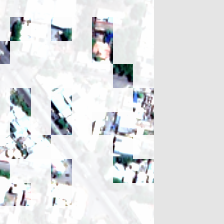}
  \end{subfigure}
  \begin{subfigure}{0.18\linewidth}
  \includegraphics[width=1\linewidth]{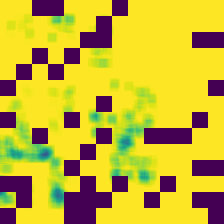}
  \end{subfigure}

  \begin{subfigure}{0.18\linewidth}
  \includegraphics[width=1\linewidth]{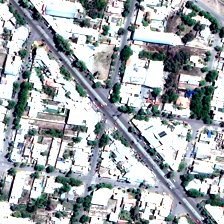}
  \end{subfigure}
  \begin{subfigure}{0.18\linewidth}
  \includegraphics[width=1\linewidth]{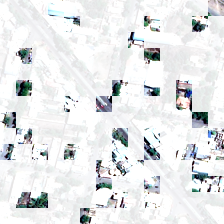}
  \end{subfigure}
  \begin{subfigure}{0.18\linewidth}
  \includegraphics[width=1\linewidth]{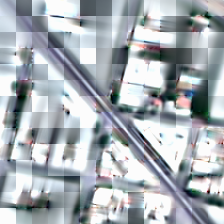}
  \end{subfigure}
  \begin{subfigure}{0.18\linewidth}
  \includegraphics[width=1\linewidth]{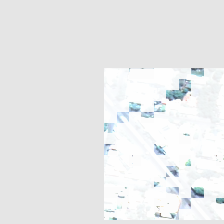}
  \end{subfigure}
  \begin{subfigure}{0.18\linewidth}
  \includegraphics[width=1\linewidth]{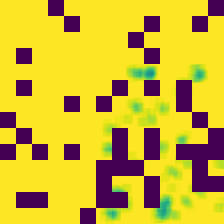}
  \end{subfigure}
  
  \vline\ 
  
  \begin{subfigure}{0.18\linewidth}
  \includegraphics[width=1\linewidth]{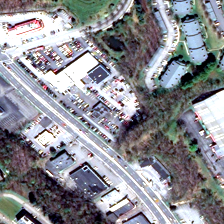}
  \end{subfigure}
  \begin{subfigure}{0.18\linewidth}
  \includegraphics[width=1\linewidth]{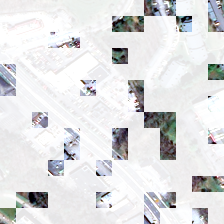}
  \end{subfigure}
  \begin{subfigure}{0.18\linewidth}
  \includegraphics[width=1\linewidth]{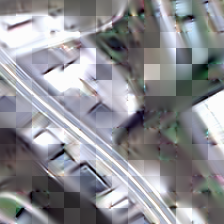}
  \end{subfigure}
  \begin{subfigure}{0.18\linewidth}
  \includegraphics[width=1\linewidth]{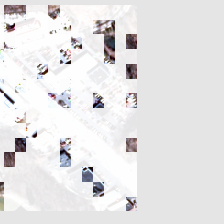}
  \end{subfigure}
  \begin{subfigure}{0.18\linewidth}
  \includegraphics[width=1\linewidth]{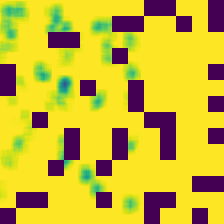}
  \end{subfigure}

  \begin{subfigure}{0.18\linewidth}
  \includegraphics[width=1\linewidth]{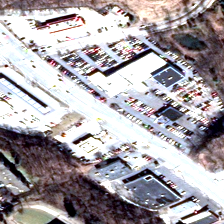}
  \end{subfigure}
  \begin{subfigure}{0.18\linewidth}
  \includegraphics[width=1\linewidth]{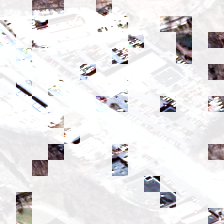}
  \end{subfigure}
  \begin{subfigure}{0.18\linewidth}
  \includegraphics[width=1\linewidth]{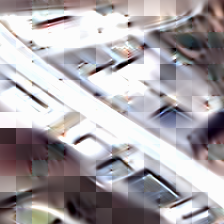}
  \end{subfigure}
  \begin{subfigure}{0.18\linewidth}
  \includegraphics[width=1\linewidth]{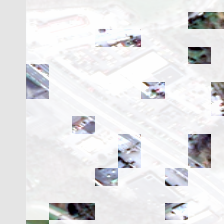}
  \end{subfigure}
  \begin{subfigure}{0.18\linewidth}
  \includegraphics[width=1\linewidth]{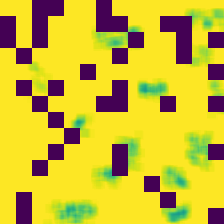}
  \end{subfigure}

  \vline\
  
  \begin{subfigure}{0.18\linewidth}
  \includegraphics[width=1\linewidth]{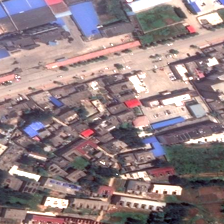}
  \end{subfigure}
  \begin{subfigure}{0.18\linewidth}
  \includegraphics[width=1\linewidth]{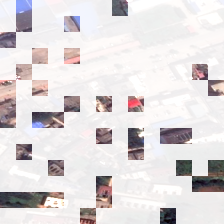}
  \end{subfigure}
  \begin{subfigure}{0.18\linewidth}
  \includegraphics[width=1\linewidth]{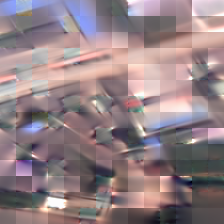}
  \end{subfigure}
  \begin{subfigure}{0.18\linewidth}
  \includegraphics[width=1\linewidth]{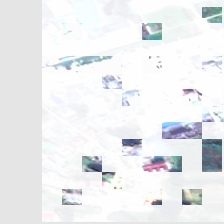}
  \end{subfigure}
  \begin{subfigure}{0.18\linewidth}
  \includegraphics[width=1\linewidth]{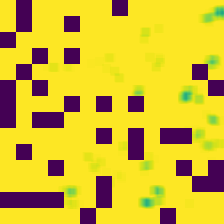}
  \end{subfigure}

  \begin{subfigure}{0.18\linewidth}
  \includegraphics[width=1\linewidth]{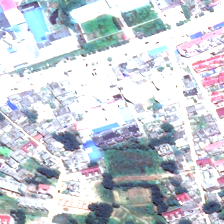}
  \end{subfigure}
  \begin{subfigure}{0.18\linewidth}
  \includegraphics[width=1\linewidth]{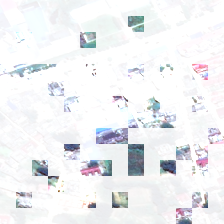}
  \end{subfigure}
  \begin{subfigure}{0.18\linewidth}
  \includegraphics[width=1\linewidth]{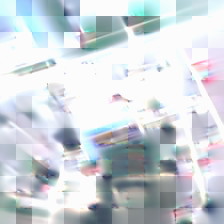}
  \end{subfigure}
  \begin{subfigure}{0.18\linewidth}
  \includegraphics[width=1\linewidth]{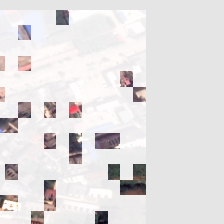}
  \end{subfigure}
  \begin{subfigure}{0.18\linewidth}
  \includegraphics[width=1\linewidth]{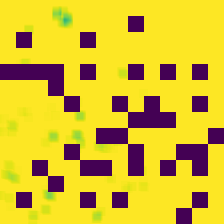}
  \end{subfigure}

  \vline\ 
  
  \begin{subfigure}{0.18\linewidth}
  \includegraphics[width=1\linewidth]{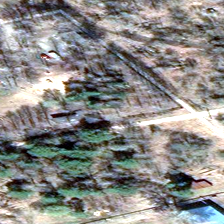}
  \end{subfigure}
  \begin{subfigure}{0.18\linewidth}
  \includegraphics[width=1\linewidth]{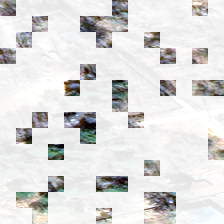}
  \end{subfigure}
  \begin{subfigure}{0.18\linewidth}
  \includegraphics[width=1\linewidth]{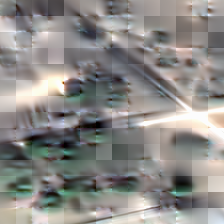}
  \end{subfigure}
  \begin{subfigure}{0.18\linewidth}
  \includegraphics[width=1\linewidth]{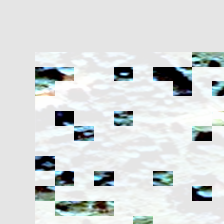}
  \end{subfigure}
  \begin{subfigure}{0.18\linewidth}
  \includegraphics[width=1\linewidth]{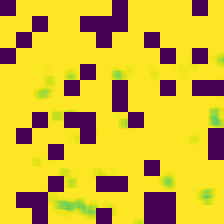}
  \end{subfigure}

  \begin{subfigure}{0.18\linewidth}
  \includegraphics[width=1\linewidth]{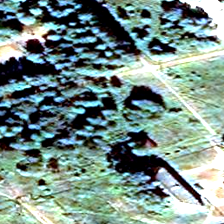}
  \caption*{Pair}
  \end{subfigure}
  \begin{subfigure}{0.18\linewidth}
  \includegraphics[width=1\linewidth]{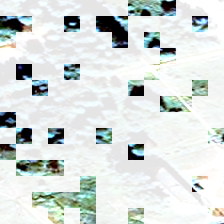}
  \caption*{Mask}
  \end{subfigure}
  \begin{subfigure}{0.18\linewidth}
  \includegraphics[width=1\linewidth]{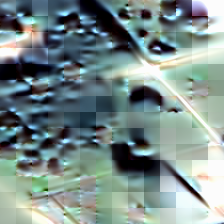}
  \caption*{Pred.}
  \end{subfigure}
  \begin{subfigure}{0.18\linewidth}
  \includegraphics[width=1\linewidth]{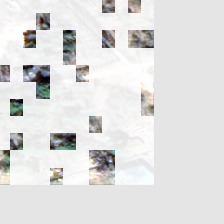}
  \caption*{Cross}
  \end{subfigure}
  \begin{subfigure}{0.18\linewidth}
  \includegraphics[width=1\linewidth]{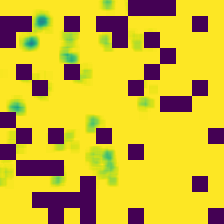}
  \caption*{Weight}
  \end{subfigure}

  \caption{Visualization of the reconstruction of neighboring images from fMoW-RGB. From left to right, we show pairs of neighboring images, masked images, prediction, cross-visible pixels, and the loss weight. Neighboring images from fMoW-RGB usually exhibit significant temporal changes and therefore our loss weighting by cross visibility has less impact.}
  \label{vis_fmow}
\end{figure}

\begin{figure}[h]
  \centering
  \begin{subfigure}{0.18\linewidth}
  \includegraphics[width=1\linewidth]{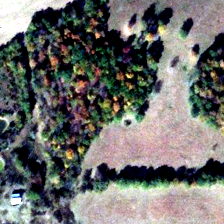}
  \end{subfigure}
  \begin{subfigure}{0.18\linewidth}
  \includegraphics[width=1\linewidth]{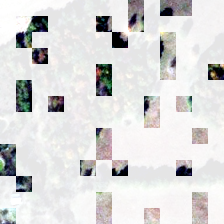}
  \end{subfigure}
  \begin{subfigure}{0.18\linewidth}
  \includegraphics[width=1\linewidth]{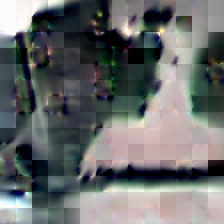}
  \end{subfigure}
  \begin{subfigure}{0.18\linewidth}
  \includegraphics[width=1\linewidth]{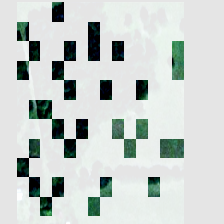}
  \end{subfigure}
  \begin{subfigure}{0.18\linewidth}
  \includegraphics[width=1\linewidth]{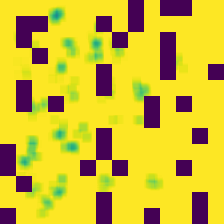}
  \end{subfigure}

  \begin{subfigure}{0.18\linewidth}
  \includegraphics[width=1\linewidth]{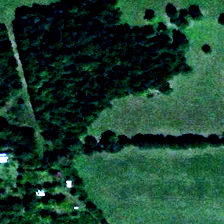}
  \end{subfigure}
  \begin{subfigure}{0.18\linewidth}
  \includegraphics[width=1\linewidth]{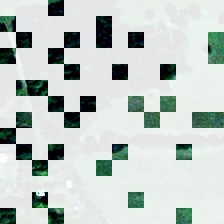}
  \end{subfigure}
  \begin{subfigure}{0.18\linewidth}
  \includegraphics[width=1\linewidth]{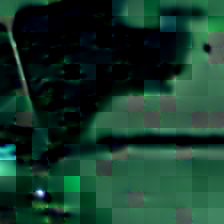}
  \end{subfigure}
  \begin{subfigure}{0.18\linewidth}
  \includegraphics[width=1\linewidth]{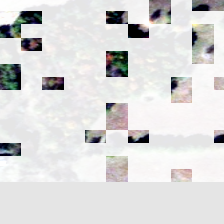}
  \end{subfigure}
  \begin{subfigure}{0.18\linewidth}
  \includegraphics[width=1\linewidth]{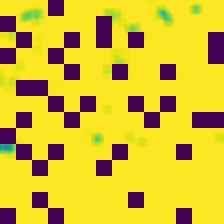}
  \end{subfigure}

  \vline\ 

  \begin{subfigure}{0.18\linewidth}
  \includegraphics[width=1\linewidth]{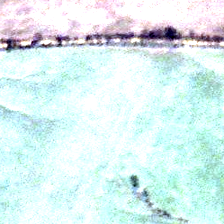}
  \end{subfigure}
  \begin{subfigure}{0.18\linewidth}
  \includegraphics[width=1\linewidth]{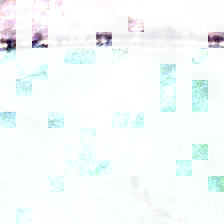}
  \end{subfigure}
  \begin{subfigure}{0.18\linewidth}
  \includegraphics[width=1\linewidth]{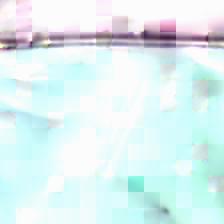}
  \end{subfigure}
  \begin{subfigure}{0.18\linewidth}
  \includegraphics[width=1\linewidth]{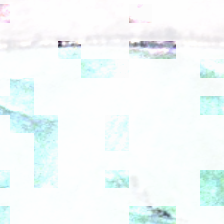}
  \end{subfigure}
  \begin{subfigure}{0.18\linewidth}
  \includegraphics[width=1\linewidth]{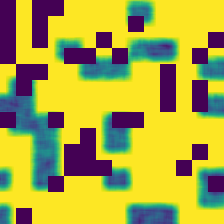}
  \end{subfigure}

  \begin{subfigure}{0.18\linewidth}
  \includegraphics[width=1\linewidth]{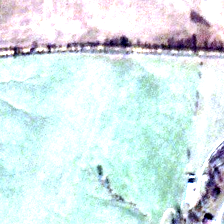}
  \end{subfigure}
  \begin{subfigure}{0.18\linewidth}
  \includegraphics[width=1\linewidth]{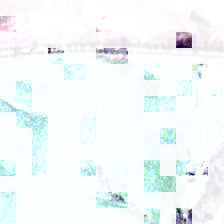}
  \end{subfigure}
  \begin{subfigure}{0.18\linewidth}
  \includegraphics[width=1\linewidth]{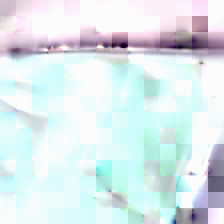}
  \end{subfigure}
  \begin{subfigure}{0.18\linewidth}
  \includegraphics[width=1\linewidth]{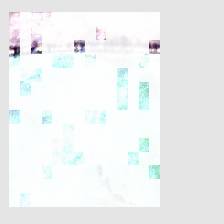}
  \end{subfigure}
  \begin{subfigure}{0.18\linewidth}
  \includegraphics[width=1\linewidth]{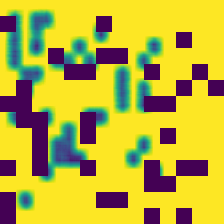}
  \end{subfigure}

  \vline\ 

  \begin{subfigure}{0.18\linewidth}
  \includegraphics[width=1\linewidth]{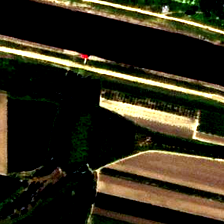}
  \end{subfigure}
  \begin{subfigure}{0.18\linewidth}
  \includegraphics[width=1\linewidth]{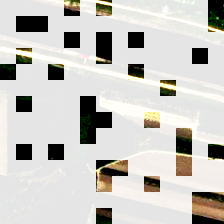}
  \end{subfigure}
  \begin{subfigure}{0.18\linewidth}
  \includegraphics[width=1\linewidth]{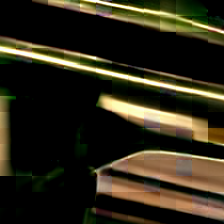}
  \end{subfigure}
  \begin{subfigure}{0.18\linewidth}
  \includegraphics[width=1\linewidth]{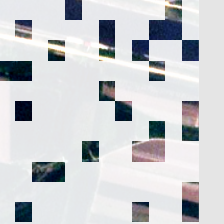}
  \end{subfigure}
  \begin{subfigure}{0.18\linewidth}
  \includegraphics[width=1\linewidth]{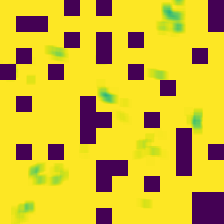}
  \end{subfigure}
  
  \begin{subfigure}{0.18\linewidth}
  \includegraphics[width=1\linewidth]{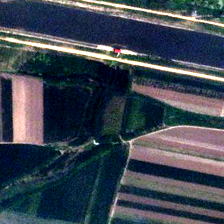}
  \end{subfigure}
  \begin{subfigure}{0.18\linewidth}
  \includegraphics[width=1\linewidth]{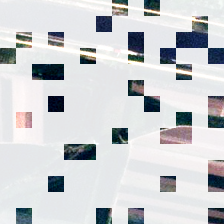}
  \end{subfigure}
  \begin{subfigure}{0.18\linewidth}
  \includegraphics[width=1\linewidth]{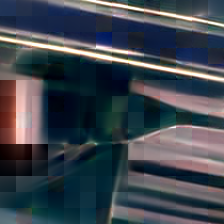}
  \end{subfigure}
  \begin{subfigure}{0.18\linewidth}
  \includegraphics[width=1\linewidth]{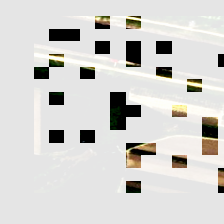}
  \end{subfigure}
  \begin{subfigure}{0.18\linewidth}
  \includegraphics[width=1\linewidth]{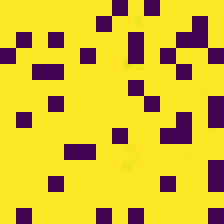}
  \end{subfigure}

   \vline\ 

  \begin{subfigure}{0.18\linewidth}
  \includegraphics[width=1\linewidth]{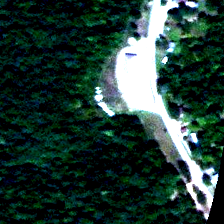}
  \end{subfigure}
  \begin{subfigure}{0.18\linewidth}
  \includegraphics[width=1\linewidth]{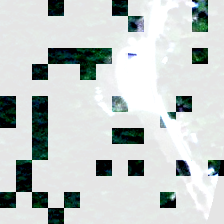}
  \end{subfigure}
  \begin{subfigure}{0.18\linewidth}
  \includegraphics[width=1\linewidth]{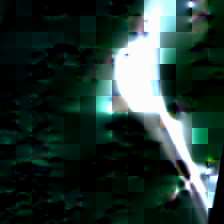}
  \end{subfigure}
  \begin{subfigure}{0.18\linewidth}
  \includegraphics[width=1\linewidth]{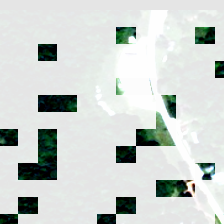}
  \end{subfigure}
  \begin{subfigure}{0.18\linewidth}
  \includegraphics[width=1\linewidth]{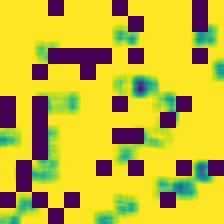}
  \end{subfigure}
  
  \begin{subfigure}{0.18\linewidth}
  \includegraphics[width=1\linewidth]{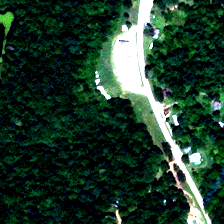}
  \caption*{Pair}
  \end{subfigure}
  \begin{subfigure}{0.18\linewidth}
  \includegraphics[width=1\linewidth]{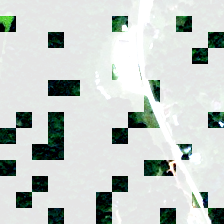}
  \caption*{Mask}
  \end{subfigure}
  \begin{subfigure}{0.18\linewidth}
  \includegraphics[width=1\linewidth]{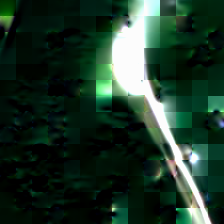}
  \caption*{Pred.}
  \end{subfigure}
  \begin{subfigure}{0.18\linewidth}
  \includegraphics[width=1\linewidth]{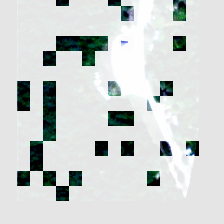}
  \caption*{Cross}
  \end{subfigure}
  \begin{subfigure}{0.18\linewidth}
  \includegraphics[width=1\linewidth]{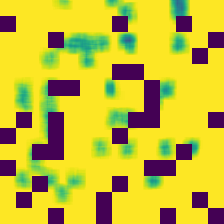}
  \caption*{Weight}
  \end{subfigure}

  \caption{Visualization of the reconstruction of neighboring images from Satellogic-RGB. From left to right, we show pairs of neighboring images, masked images, prediction, cross-visible pixels, and the loss weight. Neighboring images from Satellogic-RGB present fewer changes (fewer and even no revisits from the data). are even identical images. Our loss weighting by input visibility are more effective on Satellogic-RGB to avoid short learning shortcuts.}
  \label{vis_sat}
\end{figure}

\subsection{Reconstruction}
\label{sec:reconstruction}

We visualize the reconstruction of neighboring images from fMoW-RGB~\cite{DBLP:conf/cvpr/ChristieFWM18}  and Satellogic-RGB~\cite{DBLP:journals/corr/abs-2501-08111} in Figure ~\ref{vis_fmow} and ~\ref{vis_sat}. fMoW-RGB shows more temporal changes and semantic contents in images and Satellogic-RGB has fewer revisits and its contents are not semantic-aware. Therefore, our loss weighting by cross visibility can alleviate the information leak when training on Satellogic-RGB and improve performance.

\subsection{Attention Map}
\label{sec:attentionmap}

\begin{figure}[h]
  \centering
  \begin{subfigure}{0.24\linewidth}
    \includegraphics[width=1\linewidth]{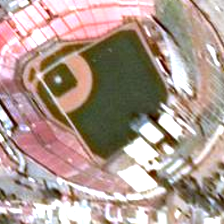}
  \end{subfigure}
  \begin{subfigure}{0.24\linewidth}
    \includegraphics[width=1\linewidth]{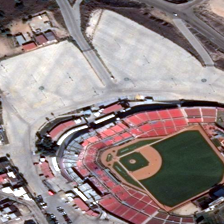}
  \end{subfigure}
  \begin{subfigure}{0.24\linewidth}
    \includegraphics[width=1\linewidth]{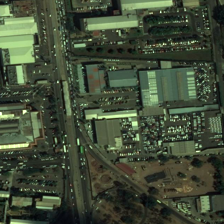}
  \end{subfigure}
  \begin{subfigure}{0.24\linewidth}
    \includegraphics[width=1\linewidth]{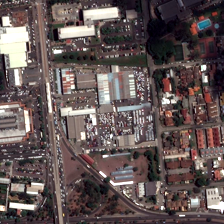}
  \end{subfigure}
  \begin{subfigure}{0.24\linewidth}
    \includegraphics[width=1\linewidth]{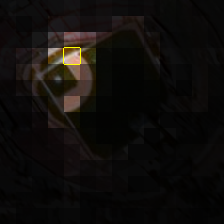}
  \end{subfigure}
  \begin{subfigure}{0.24\linewidth}
    \includegraphics[width=1\linewidth]{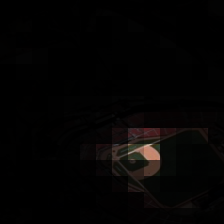}
  \end{subfigure}
    \begin{subfigure}{0.24\linewidth}
    \includegraphics[width=1\linewidth]{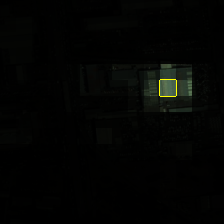}
  \end{subfigure}
  \begin{subfigure}{0.24\linewidth}
    \includegraphics[width=1\linewidth]{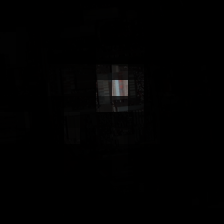}
  \end{subfigure}
  \begin{subfigure}{0.24\linewidth}
    \includegraphics[width=1\linewidth]{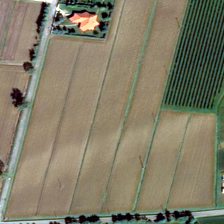}
  \end{subfigure}
  \begin{subfigure}{0.24\linewidth}
    \includegraphics[width=1\linewidth]{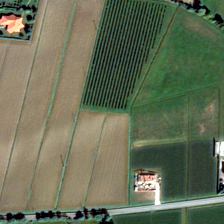}
  \end{subfigure}
  \begin{subfigure}{0.24\linewidth}
    \includegraphics[width=1\linewidth]{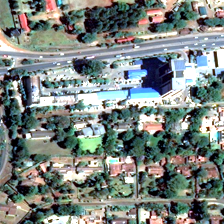}
  \end{subfigure}
  \begin{subfigure}{0.24\linewidth}
    \includegraphics[width=1\linewidth]{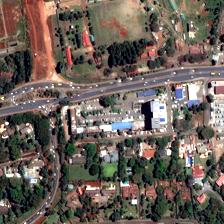}
  \end{subfigure}
  \begin{subfigure}{0.24\linewidth}
    \includegraphics[width=1\linewidth]{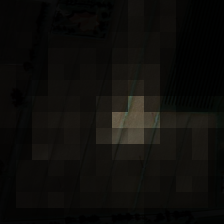}
  \end{subfigure}
  \begin{subfigure}{0.24\linewidth}
    \includegraphics[width=1\linewidth]{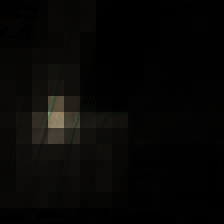}
  \end{subfigure}
  \begin{subfigure}{0.24\linewidth}
    \includegraphics[width=1\linewidth]{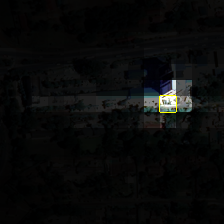}
  \end{subfigure}
  \begin{subfigure}{0.24\linewidth}
    \includegraphics[width=1\linewidth]{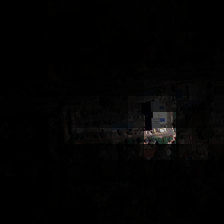}
  \end{subfigure}
  \caption{Visualization of the attention of neighboring images. The first row shows the input neighboring images and the second row shows the attention scores with respect the to the patch marked by yellow boundary in the first image. The visualization indicates that NeighborMAE can build spatial relations across neighboring images and learn correlated representations.}
  \label{attn}
\end{figure}

We show the attention map associated with a specific patch when unmasked neighboring images are fed to the learned models in ~\ref{attn}. The corresponding regions from neighboring images receive high attention scores, which indicates that NeighborMAE can learn spatial dependency from neighboring images.

\subsection{Spatial and Temporal Distributions of Datasets}

\begin{figure}[h]
  \centering
  \begin{subfigure}{0.24\linewidth}
    \includegraphics[width=1\linewidth]{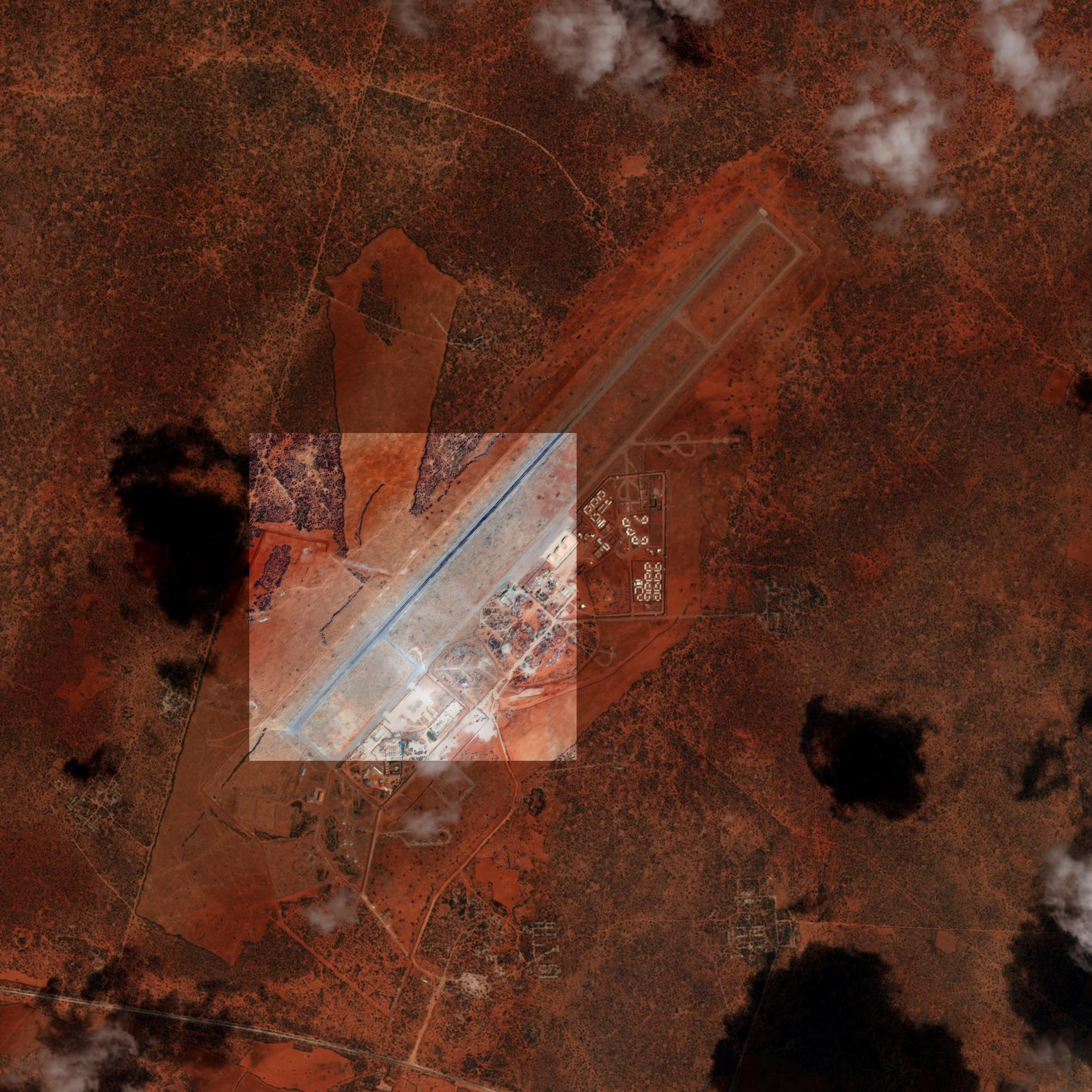}
  \end{subfigure}
  \begin{subfigure}{0.24\linewidth}
    \includegraphics[width=1\linewidth]{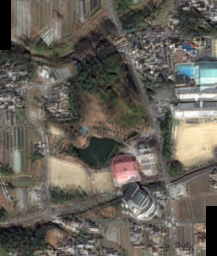}
  \end{subfigure}
  \begin{subfigure}{0.24\linewidth}
    \includegraphics[width=1\linewidth]{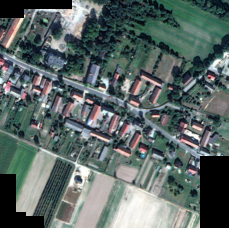}
  \end{subfigure}
  \begin{subfigure}{0.24\linewidth}
    \includegraphics[width=1\linewidth]{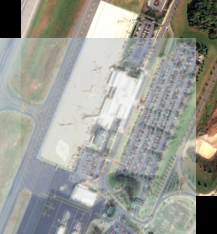}
  \end{subfigure}
  \caption{We show spatial and temporal distributions of fMoW. Pixel values are averaged over overlapping regions. The image sizes are mostly different in fMoW and the images represent specific areas of interest. fMoW features organized temporal sequences captured by multiple EO projects.}
  \label{fmow_st}
\end{figure}

\begin{figure}[h]
  \centering
  \begin{subfigure}{0.24\linewidth}
    \includegraphics[width=1\linewidth]{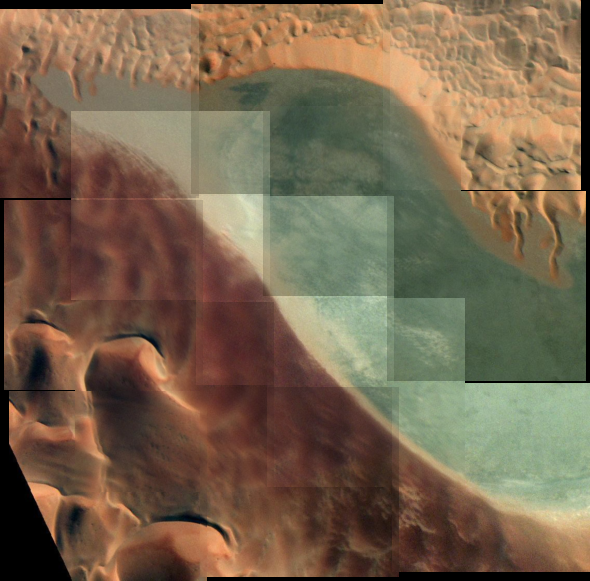}
  \end{subfigure}
  \begin{subfigure}{0.24\linewidth}
    \includegraphics[width=1\linewidth]{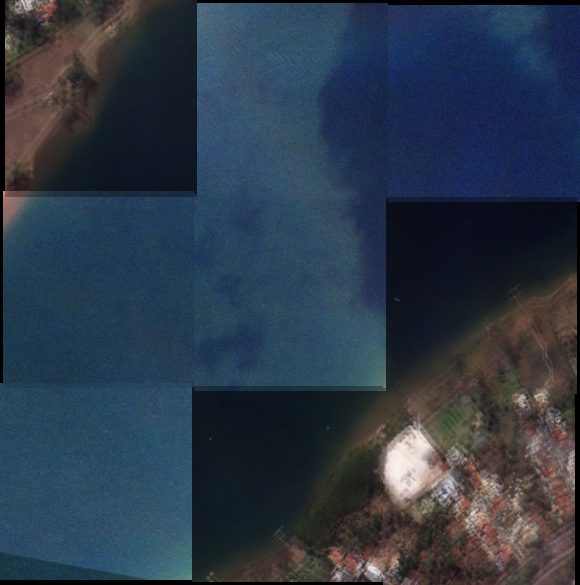}
  \end{subfigure}
  \begin{subfigure}{0.24\linewidth}
    \includegraphics[width=1\linewidth]{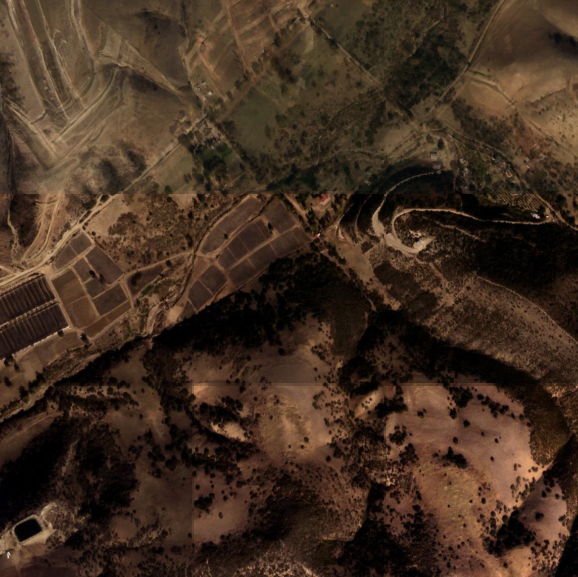}
  \end{subfigure}
  \begin{subfigure}{0.24\linewidth}
    \includegraphics[width=1\linewidth]{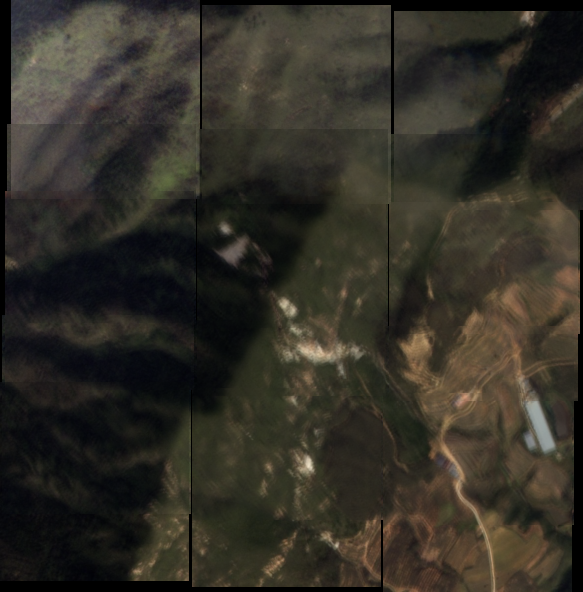}
  \end{subfigure}
  \caption{We show spatial and temporal distributions of Satellogic. Pixel values are averaged over overlapping regions. The image sizes are consistent in Satellogic and the image contents are not semantic-aware. Few or no revisits are provided for specific locations.}
  \label{satl_st}
\end{figure}

Examples of images from the used fMoW~\cite{DBLP:conf/cvpr/ChristieFWM18} and Satellogic~\cite{DBLP:journals/corr/abs-2501-08111} datasets for pretraining are displayed in Figure ~\ref{fmow_st} and Figure ~\ref{satl_st}. fMoW and Satellogic present different spatial and temporal characteristics, and NeighborMAE is robust and achieves significant improvement on both datasets.

\section{Statistical Significance}
To assess whether the performance improvement by NeighborMAE is statistically significant, we perform 5 independent runs of consecutive SSL pretraining and evaluation on fMoW, following the same setting as the main experiment in ~\ref{main result}. From table ~\ref{stat}, we can see that the standard deviations are relatively small for both methods, and the mean difference between the baseline MAE and our NeighborMAE is large enough. This suggests that the observed performance gain is unlikely to be due to random variation and can be attributed to the effectiveness of our method.

\begin{table}[h]
  \scriptsize
  \centering
  \begin{tabular}{@{}c|c|c|c|c|c|c @{}}
    \toprule
    Method   &   $1^{st}$   &   $2^{nd}$    &   $3^{rd}$   &  $4^{th}$   &  $5^{th}$  & mean$\pm$std  \\
    \midrule
    MAE         & 67.19   &  66.99    &   66.71      &   66.56      &   65.99      &  66.69$\pm$0.41  \\
    NeighborMAE & 69.42   & 69.33   &  68.90    & 68.61      & 68.08    &  68.87$\pm$0.49   \\
    \midrule
    MAE         &  78.39  &  78.22    &   78.13     &   78.07     &   77.91    &  78.14$\pm$0.16   \\
    NeighborMAE &  79.53  & 79.53  & 79.26     & 79.24    &  79.11   &   79.33$\pm$0.17   \\
        
    \bottomrule
  \end{tabular}
  \caption{Performance of 5 independent runs of consecutive SSL pretraining and evaluation on fMoW, using baseline MAE and our NeighborMAE with the main experiment setting in ~\ref{main result}. The upper and lower part shows the accuracy of linear probing and fine-tuning, respectively. We rank the performance from left to right. The results demonstrate that the improvement achieved by NeighborMAE is beyond the deviation caused by randomness.}
  \label{stat}
\end{table}

\end{document}